\documentclass{article} 
\usepackage{iclr-2027-style-files/iclr2027/iclr2027_conference}
\usepackage{times}
\usepackage{comment}
\usepackage{graphicx}
\usepackage{wrapfig}
\usepackage{caption}
\usepackage{algorithm}
\usepackage{algpseudocode}
\algrenewcommand{\algorithmiccomment}[1]{\hfill\textit{// #1}}
\usepackage{enumerate}
\usepackage{enumitem} 
\usepackage{amsmath}
\usepackage{bm}
\usepackage{booktabs,makecell,multirow}
\usepackage{xcolor}
\usepackage{colortbl}
\usepackage{capt-of}
\usepackage[font=small]{caption}
\usepackage{float}
\usepackage{tabularx}

\usepackage{amsmath,amsfonts,bm}

\def\eqref#1{equation~\ref{#1}}

\def\1{\bm{1}}

\DeclareMathAlphabet{\mathsfit}{\encodingdefault}{\sfdefault}{m}{sl}
\SetMathAlphabet{\mathsfit}{bold}{\encodingdefault}{\sfdefault}{bx}{n}

\newcommand{\R}{\mathbb{R}}

\usepackage{hyperref}
\usepackage{url}

\title{Accelerating VLA Inference with Perturbation Informed Execution}

\author{%
  Jehun Lee, Hajun Lee, Jimin Ryu, Youhan Lee, Beomseok Kang, \\
  \textbf{Jonghyun Choi, Jae-Joon Kim}\\[0.2em]
  Seoul National University\\[0.2em]
  \texttt{\{jehun.lee, kimjaejoon\}@snu.ac.kr}%
}

\iclrarxivcopy  
\begin{document}

\maketitle
\begin{abstract}
Action-chunking vision-language-action (VLA) models predict sequences of future robot actions.
Under fixed-horizon execution, the robot executes a fixed-length prefix of each predicted chunk before invoking the policy again, leaving the remaining predictions unused. 
For a given number of executed actions, short execution horizons require more policy calls.
Extending the execution horizon amortizes policy inference cost over more actions. 
However, uniformly applying long horizons across episodes can limit opportunities for precise corrections.
To address this limitation, 
we propose PICES (Perturbation-Informed Chunk and Execution Horizon Selection), a training-free method that jointly selects an action chunk and the number of actions executed before the next policy call. 
Within a single policy call, the action expert generates a batch of action chunks from multiple initial noise samples and their locally perturbed counterparts, using the same conditioning features across all samples.
PICES exploits differences between corresponding actions in each nominal chunk and its counterpart to determine candidate-specific execution horizons and select a nominal chunk for execution. 
Experiments on simulation benchmarks and real-world manipulation tasks demonstrate that PICES achieves speedup by up to 2.3$\times$ in wall-clock policy inference time on Jetson Orin and Thor relative to fixed-horizon baselines while improving the success rate. 

\end{abstract}

\section{Introduction}

\begin{figure}[t]
    \centering
    \includegraphics[width=\linewidth]{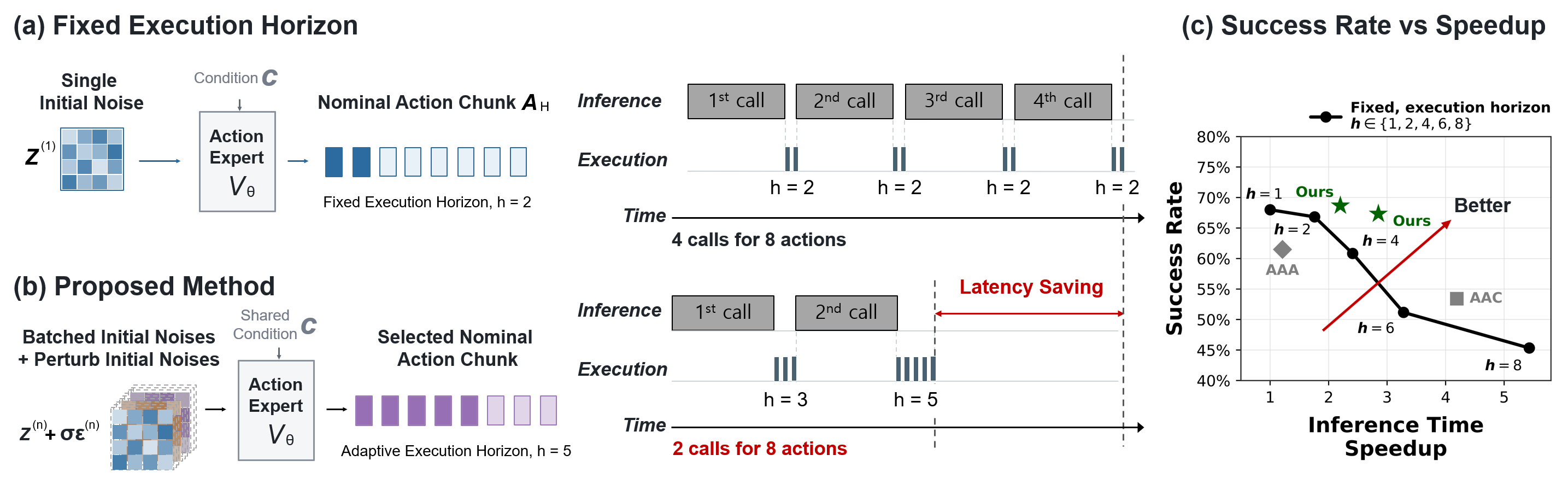}
    \vspace{-3mm}
    \caption{\textbf{High-level idea of the PICES.} (a) Fixed execution horizon generates a nominal action chunk from a single initial noise sample at each policy call and executes a fixed-length actions before replanning. 
    (b) PICES uses multiple initial-noise samples and their perturbations to determine execution horizon of nominal action chunk. 
    Our method reduces the number of policy calls required to complete the same episode, from four to two in the illustrated example. 
    (c) Success rate versus speedup comparison across methods, showing that PICES achieves wall-clock inference time speedup while preserving success rate.}
    \label{fig:Figure1}
    \vspace{-3mm}
\end{figure}

Vision-language-action (VLA) models generate robot actions conditioned on visual observations, proprioceptive states, and language instructions~\citep{brohan2022rt, kim2024openvla, kim2025fine}.
Recent VLA architectures pair a vision-language backbone with a flow-matching action expert that transforms initial noise into an action chunk, a sequence of future actions~\citep{bjorck2025gr00t, black2024pi_0, intelligence2025pi_, shukor2025smolvla}. 
In conventional action chunking, the robot executes fixed-length actions of each predicted chunk before replanning as illustrated in Figure~\ref{fig:Figure1}(a). 
The number of actions executed before update observations defines the execution horizon~\citep{black2026real}.
The cumulative policy invocations across policy rollout makes task-completion time dependent on both per-call latency and the number of policy calls~\citep{zhou2026characterizing, jing2025mixture}.

At comparable success rates, shorter task-completion times enable a robot to complete more tasks within a fixed time budget~\citep{shukor2025smolvla, huang2025dadu}.
This makes cumulative inference latency as an important consideration for onboard VLA deployment, where edge platforms such as NVIDIA Jetson Orin and Thor operate under compute and memory-bandwidth constraints~\citep{yang2026jetson,jiang2026fast}.
For example, GR00T N1.7 requires 80.4ms per policy call on Jetson Orin and 150.9 ms on Jetson Thor~\citep{ FluxVLA2026,bjorck2025gr00t}, whereas observation acquisition and actuation can operate up to 60Hz and 1kHz, respectively~\citep{jiang2026fast, ma2025running, zhang2020modular}.
In observation-inference-execution pipeline, reducing the number of policy calls lowers cumulative inference latency and shorten task-completion time~\citep{jing2025mixture}.
For action chunking policies, executing only fraction of each predicted action chunk incurs inefficiency. 
However, applying long horizons uniformly across task phases delays the incorporation of updated observations when precise control is required.
This motivates adapting execution horizons to reduce repeated inference while preserving task success.

Prior works have explored diverse action-chunk execution strategies to balance temporal consistency and reactivity, including temporal ensembling, adaptive chunk updates, and guided candidate selection~\citep{zhao2023learning,so2026improving,liu2025bidirectional}. 
AAC estimates action entropy of action candidates to decide a shared exceuction horizon~\citep{liang2026adaptive}.
AutoHorizon derives the execution horizon from self-attention within a predicted chunk~\citep{wang2026vla}.
\text{$A^{3}$} selects a representative trajectory from multiple candidates and then performs additional flow-matching verification to determine execution horizon~\citep{zhao2026dynamic}. 
Our objective is to reduce wall-clock inference time while achieving comparable success rate by jointly selecting an action chunk and its execution horizon.
This raises our key question: \textbf{Given a bounded set of candidate action chunks generated in a single policy call, how can we select the action chunk and execution horizon to reduce the wall-clock inference time?}

We propose PICES (Perturbation-Informed Chunk and Execution Horizon Selection), a training-free inference time method that jointly selects an action chunk and its execution horizon. 
For each initial-noise sample, PICES constructs a locally perturbed counterpart by adding a small Gaussian perturbation. 
A single batched policy call generates the corresponding nominal and auxiliary action chunks using shared conditioning features encoded by the vision-language-model (VLM).
Based on the arithmetic intensity analysis of action expert and hardware ridge-point in our motivational analysis, we restrict total batch size to a range with negligible incremental latency.
For each candidate, the stepwise differences between nominal and auxiliary action chunks yield a local sensitivity value.
We exploit these values to retain candidates with low initial sensitivity and estimate execution horizon for each candidate. 
From the candidate pool, PICES selects and executes the nominal chunk with the longest execution horizon.
Figure~\ref{fig:Figure1}(b) illustrates the reduction in policy calls: two calls execute eight actions, compared with four calls with fixed horizon execution. 
Savings from fewer policy calls outweigh the additional cost of candidate generation, reducing cumulative inference time as shown in Figure~\ref{fig:Figure1}(c).

Our key contributions are as follows:
\begin{itemize}
\item 
    Through an open-loop study, we show that oracle selection from multiple action chunks can yield lower reference-action error over a longer execution horizon than a single-candidate, short-horizon baseline. We also characterize the cost of generating candidates in a batch.

\item 
    We propose PICES, a training-free method to reduce cumulative wall-clock inference time. PICES uses sensitivity to initial-noise perturbations to guide the joint selection of an action chunk and its execution horizon.

\item 
    We evaluate PICES across multiple VLA models in simulation and on real-world robot tasks. Compared with fixed-horizon baselines, PICES reduces policy calls by 25.4--60.3\% and achieves 1.2--2.3$\times$ speedups in wall-clock inference time on Jetson Orin and Thor, while maintaining higher success rates. 

\end{itemize}

\section{Related Work}
\paragraph{Action Chunking in VLA Models.}
Recent VLAs such as CogACT, $\pi$, and GR00T series execute only a fixed-length horizon of each predicted chunk~\citep{li2024cogact,black2024pi_0,bjorck2025gr00t}. 
Executing a long fixed-length actions reduces policy calls but limits the incorporation of new observations, motivating adaptive execution horizons~\citep{liang2026adaptive, wang2026vla, chen2026dynamic} (see Appendix~\ref{app:related_action_chunk} for full discussion).
In contrast, our method instead estimates a horizon for each candidate and uses the estimated execution length to choose the reliable action chunk to shorten cumulative inference time without training.

\paragraph{Acceleration for VLA Models.}
Efficient VLA inference seeks to reduce task completion time using various methods.
To reduce inference overhead, previous work exploits redundancy in visual inputs and model computation through token pruning, channel pruning, and model scheduling~\citep{chen2024image, shukor2025smolvla, xu2026vla, feng2026see} (see Appendix ~\ref{app:related_accel} for full discussion).
Our method is orthogonal to token pruning, which reduces computation within individual policy calls. 
PICES examines how candidate-specific horizon selection can reduce policy calls, while explicitly accounting for the additional cost of generating and assessing multiple action candidates.

\section{Proposed Approach} 
\subsection{Background}

\textbf{Input Condition and Action Chunk.}
Let $\bm{c} := \bigl(\bm{o},\bm{q},\ell\bigr) \in \mathcal{C}$ denote the conditioning input, comprising the current observation $\bm{o}$, the proprioceptive state $\bm{q}$, and the language instruction $\ell$.
Let $V_\theta$ denote a pretrained VLA action generator parameterized by $\theta$, $V_\theta: \mathcal{C}\times\R^{d\times H} \longrightarrow \R^{d\times H}.$
Here, $d$ is the dimension of each action and $H$ is the number of action steps generated in one action chunk. Given an initial noise sample $\bm{Z}\in\R^{d\times H}$, the model generates
$\bm{A} = \left[ \bm{a}_0, \bm{a}_{1}, \ldots, \bm{a}_{H-1} \right] \in\R^{d\times H}$
To generate $N$ candidate action chunks from the same current input, sample $N$ independent initial noise matrices
$\bm{Z}^{(n)} \in\R^{d\times H}$. 
The $n$ indicates the candidate index. All candidates share the same conditioning prefix $\bm{c}$ and differ only in their initial noise samples:
\begin{equation}
    \bm{A}^{(n)} = V_\theta\!\left(\bm{c},\bm{Z}^{(n)}\right)
    = \left[ \bm{a}^{(n)}_0, \bm{a}^{(n)}_{1}, \ldots, \bm{a}^{(n)}_{H-1} \right]
    \in\R^{d\times H}, \qquad n=1,\ldots,N.
\end{equation}
Equivalently, conditioned on the same current visual observation, proprioceptive state, and language instruction, $N$ independent source samples are mapped into $N$ candidate action chunks with $H$ number of actions in single action chunk, whereas $h\in\{1,\ldots,H\}$ denotes the execution horizon.

\subsection{Motivational Study}
We analyze whether sampling multiple action chunks under identical conditioning inputs can provide suitable candidates for longer execution horizon than fixed-length execution. 
To limit the latency overhead of this additional sampling, we explore affordable batch size during chunk generation. 
This motivates two questions: (1) Can multiple sampled chunks provide a lower reference-action error at a longer fixed execution horizon candidate than a single sampled chunk? (2) Under consideration of action expert's arithmetic intensity, how many candidate chunks can we expect to be generated in a batch with modest additional latency?

\paragraph{Selection by minimum reference-action RMSE.}
We first compare reference action root-mean-square-error (RMSE) across fixed execution horizons and candidate counts.
We conduct the analysis with the RT-1 Fractal dataset and GR00T N1.7~\citep{bjorck2025gr00t, brohan2022rt}. 
For each condition $\bm {c}$, we generate $N\in{1,\ldots,10}$ candidate action chunks using independent initial noise and evaluate horizons $h\in\{1,\ldots,8\}$.
For each input and $(N,h)$ pair, an oracle selects the candidate with the lowest RMSE over its first $h$ actions relative to the corresponding reference actions.
The RMSE metric is explained in Appendix~\ref{app:RMSE}.
Figure~\ref{fig:oracle_rmse} shows that selection from four candidates at an eight-step horizon ($N=4$, $h=8$) achieves an RMSE of $0.1795$, compared with $0.1900$ for the single-candidate action chunk at one-step horizon ($N=1$, $h=1$).
This result demonstrates that chunk selection from multiple candidates can yield lower reference-action error even when evaluated over a longer execution horizon, motivating an inference-time method for joint action chunk and execution horizon selection.

\begin{figure}[t]
    \centering
    \begin{minipage}[t]{0.48\linewidth}
        \vspace{0pt}
        \centering
        \vspace{-1.5mm}
        \includegraphics[width=\linewidth]{
            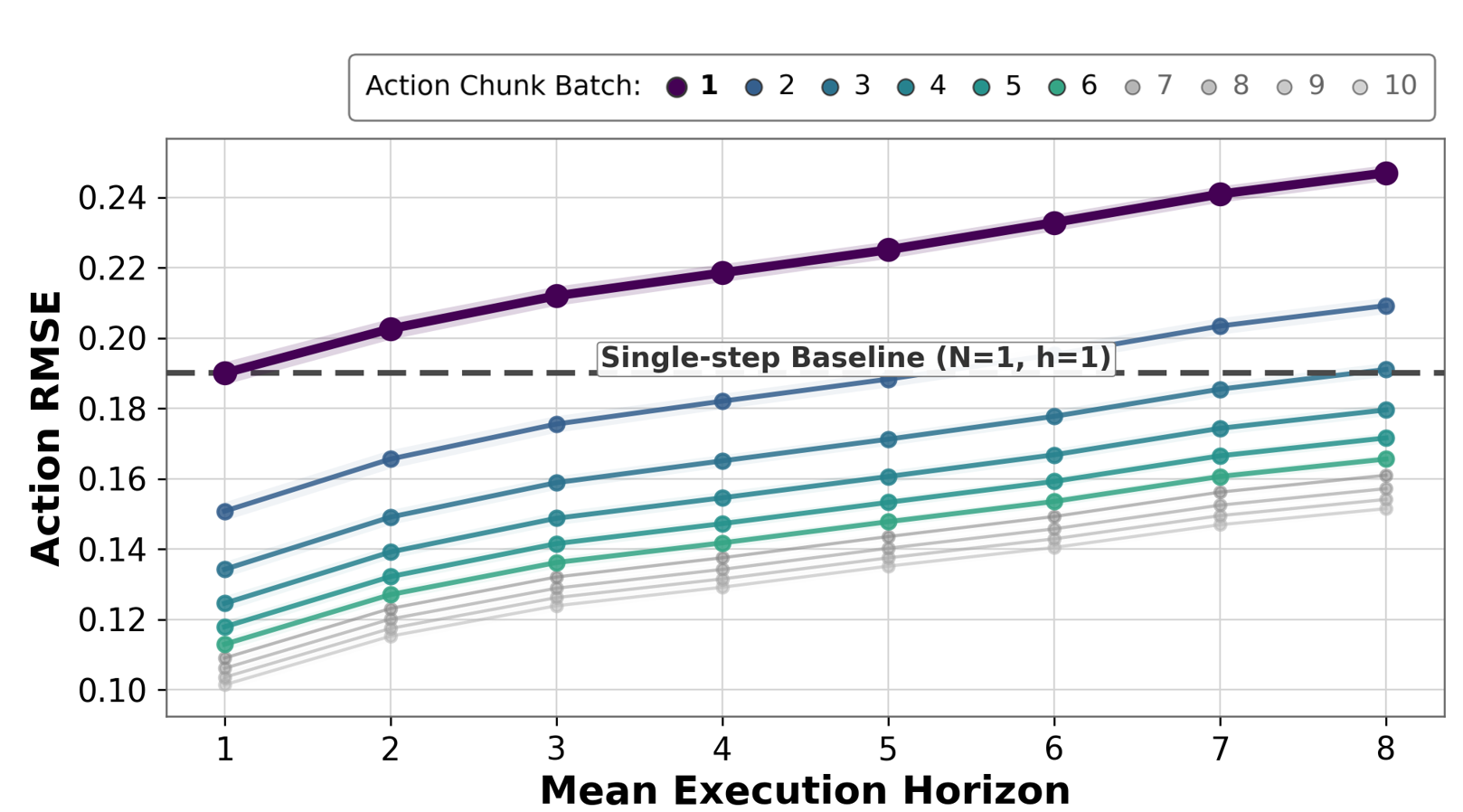
        }
        \caption{
            \textbf{
            Root-mean-square-error (RMSE) between reference and oracle-selected action chunk.}
            The dashed horizontal line marks the baseline single-candidate, one-step baselin ($N=1$, $h=1$). 
        }
        \label{fig:oracle_rmse}
    \end{minipage}\hfill%
    \begin{minipage}[t]{0.48\linewidth}
        \vspace{0pt}
        \centering
        \vspace{-2mm}
        \includegraphics[width=\linewidth]{
            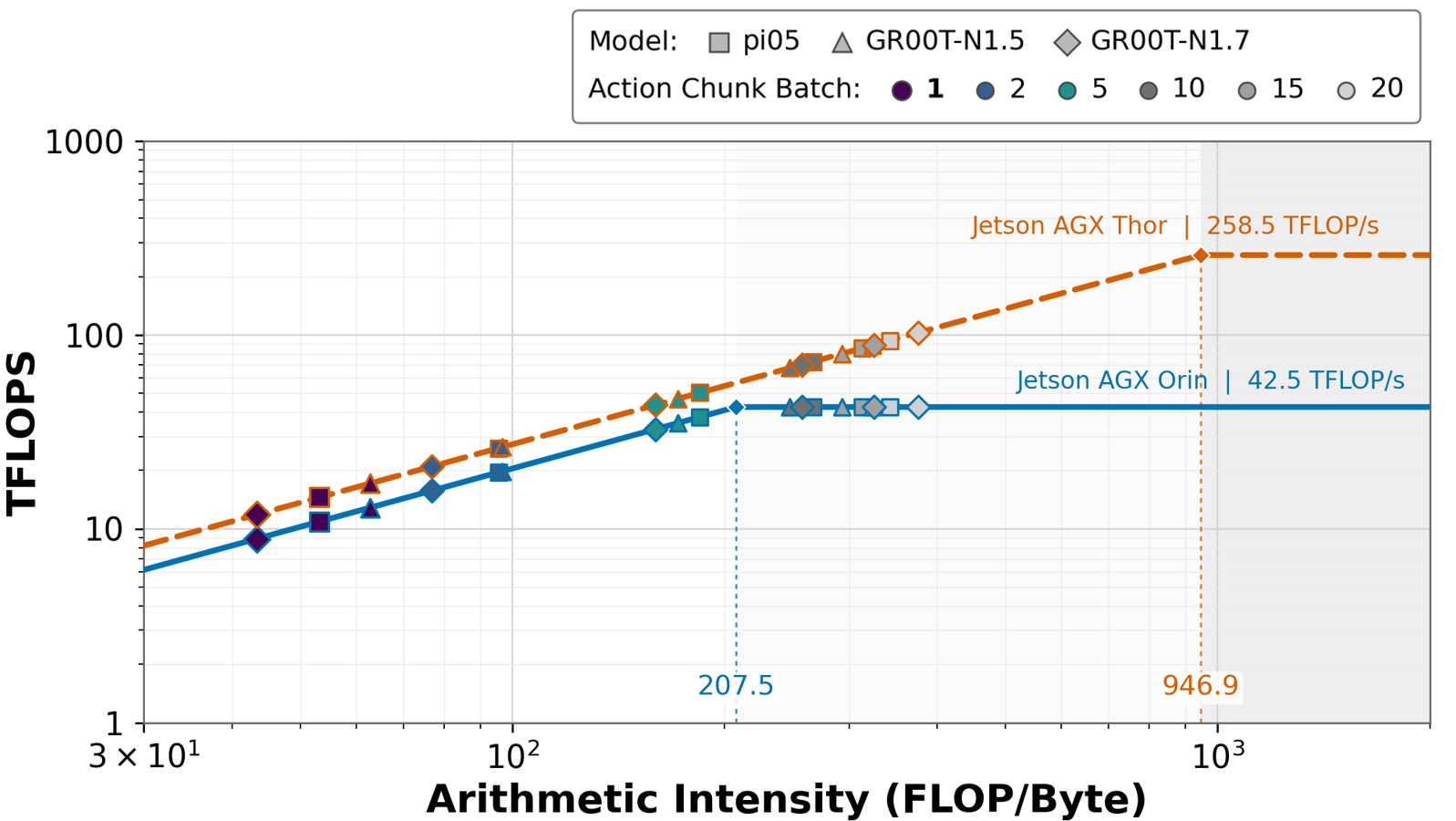
        }
        \caption{
            \textbf{Roofline analysis of batching action generation on Jetson Orin and Thor.} For each device, arithmetic intensities over its ridge point lie in the compute-bound region, where compute throughput determines the latency.
        }
        \label{fig:arithmetic_intensity}
    \end{minipage}
    \vspace{-4mm}
\end{figure}

\paragraph{Operation Characteristics of Action Expert.}
Cumulative inference-time saving for the longer execution horizons requires generating action chunks with limited additional latency. 
Prior studies identify low arithmetic intensity (AI) of action generation in action expert~\citep{ jiang2026fast, zhou2026characterizing}. 
The roofline in Figure~\ref{fig:arithmetic_intensity} shows the AI of the $\pi_{0.5}$, GR00T N1.5, and GR00T N1.7 action experts at batch sizes $B \in \{1,2,5,10,15,20\}$.
At single batch, the AI of all three action experts lies below the ridge points of both Jetson Orin and Thor, indicating memory-bound operation under the roofline model.
When AI remains below the device’s ridge point, the operation latency depends primarily on memory transfer latency. 
In such a case, the additional computation can be accommodated without a proportional increase in latency. 
On Orin, all three models approach the ridge point at a batch size of 6. 
On Thor, the action experts remain memory-bound even at a batch size of 20; however, beyond a batch size of 10, growing activation traffic incurs latency overhead. 
We therefore set the total generation batch-size limits to 6 on Orin and 10 on Thor to balance candidate diversity against per-call latency. 
Appendix~\ref{app:AI_of_action} provides the further analysis of arithmetic intensity and measured latency on GPUs.

Together, these analyses suggest that a bounded set of action chunks can contain candidates suitable for longer execution horizons with modest cost for the batched generation of candidates. 
This raises the central question: how can we assess the reliability of each candidate’s future action steps and jointly select a chunk and its execution horizon?

\subsection{Perturbation-Informed Candidate Assessment}
For each nominal action chunk, we measure changes in the action expert output under local perturbations of the corresponding initial noise.
Local sensitivity quantifies output variation under small input changes, and exploiting this sensitivity has been used to encourage invariant representations~\citep{rifai2011contractive}. 
In image generation, per-image perceptual path length quantifies how much the generated image changes perceptually under small perturbations of its latent code~\citep{karras2020analyzing}.
Motivated by these observations, we use local stability under initial-noise perturbations as a proxy for reliability of each nominal action. 
We estimate sensitivity at each action timestep within a chunk and use these estimates to favor nominal candidates with lower sensitivity.

\paragraph{Step 1 \& 2. Local perturbations of the initial noise and multiple action chunks.}
Comparing generator outputs at neighboring latent inputs provides a practical way to probe local sensitivity~\citep{odena2018generator}. 
Related analyses of diffusion sampling connect initial-noise perturbations to changes in the final generated output~\citep{song2026ccs}. 
These analyses motivate our use of initial-noise perturbations to assess the local stability of nominal action chunks.
For each nominal candidate $\bm A_t^{(n)} = V_\theta(\bm c,\bm Z^{(n)})$, we sample perturbation matrices $\bm\epsilon^{(n,r)} \in \mathbb{R}^{d\times H}$ with i.i.d. $\mathcal N(0,1)$ entries. 
All perturbation matrices are mutually independent and independent of the nominal initial noises.
\begin{equation}
\label{eq:perturbed_candidates}
\widetilde{\bm Z}^{(n)}
=\bm Z^{(n)}+\sigma\bm\epsilon^{(n)},
\qquad
\widetilde{\bm A}_t^{(n)}
=V_\theta\!\left(\bm c,\widetilde{\bm Z}^{(n)}\right),
\end{equation}
where $\sigma>0$ controls the perturbation scale.
Perturbations are applied only to the initial noise before flow integration.
We stack all nominal initial noises $\bm Z^{(n)}$ and their perturbed counterparts $\widetilde{\bm Z}^{(n)}$ into a single batch. 
All nominal and perturbed initial-noise samples are processed in a single batched policy call with a shared conditioning input $\bm c$.
For each candidate $n$, we estimate sensitivity at each action timestep by comparing the nominal chunk $\bm A^{(n)}$ with its auxiliary chunk $\widetilde{\bm A}^{(n)}$.
Only actions from the unmodified nominal chunks are used for execution.

\begin{figure}[t]
    \centering
    \includegraphics[width=\linewidth]{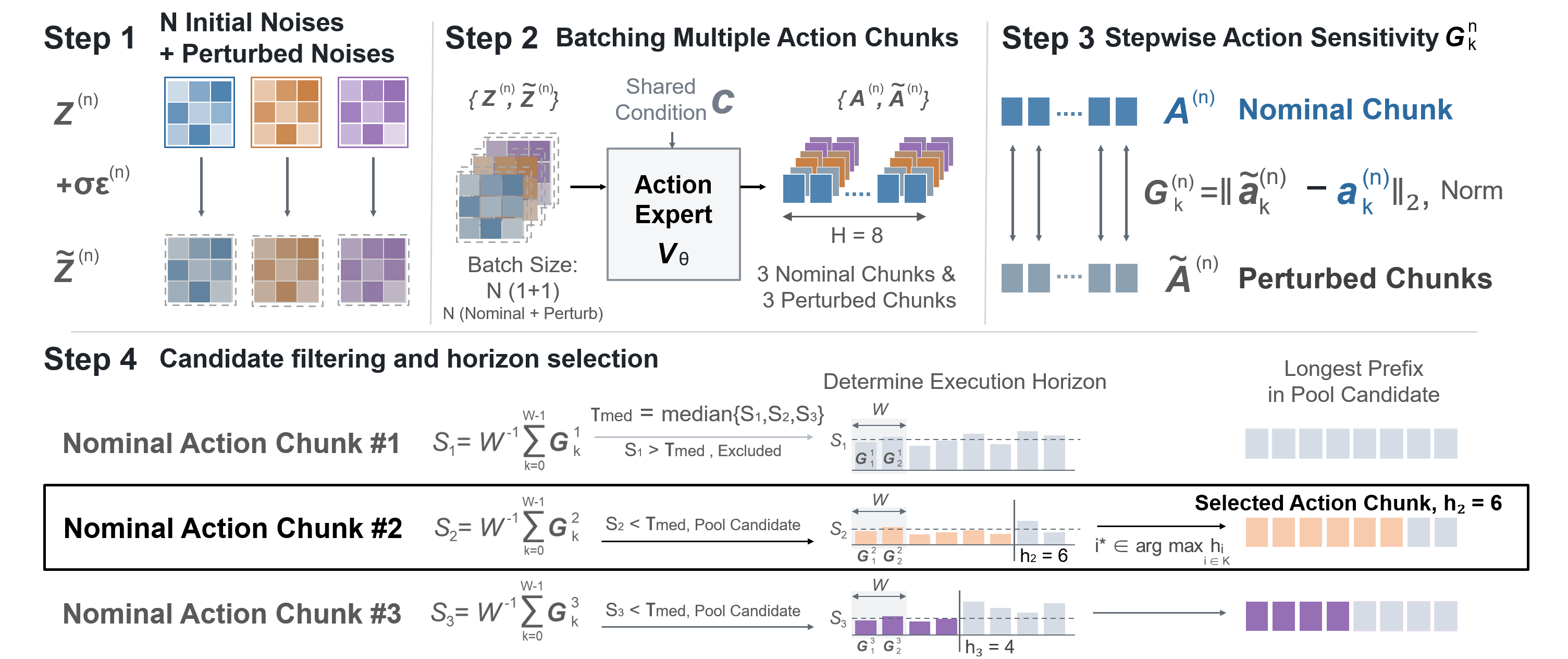}
    \caption{\textbf{Overview of PICES.} \textbf{Step 1:} Draw \(N\) initial noise samples and construct a perturbed variant of each sample. 
    \textbf{Step 2:} The action expert processes the resulting \(N(1+1)\) inputs in a single batch under the shared condition \(C\), producing \(N\) nominal action chunks and their corresponding perturbed chunks. 
    \textbf{Step 3:} Stepwise action sensitivity \(G_k^{(n)}\) is computed by comparing each nominal chunk with its perturbed counterparts. 
    \textbf{Step 4:} Compute a sensitivity score \(S_i\) for each nominal chunk by averaging its sensitivity over the first \(W\) steps and use this score to determine its execution horizon \(h_i\). 
    PICES retains candidates whose scores are at or below the median and selects the nominal chunk with the longest execution horizon from the pool.
}
    \label{fig:method_overview}
\end{figure}

\paragraph{Step 3. Stepwise action sensitivity score.}
Write the nominal action chunk as $\bm A^{(n)}=[\bm a_0^{(n)},\ldots,\bm a_{H-1}^{(n)}] \in\mathbb R^{d\times H}$, where $\bm a_k^{(n)}\in\mathbb R^d$ is the action at offset $k\in\{0,\ldots,H-1\}$. 
Define $\widetilde{\bm a}_k^{(n)}$ analogously from the auxiliary chunk $\widetilde{\bm A}^{(n)}$. 
The execution horizon $h$ denotes the number of leading actions executed from the selected chunk.
With the conditioning input $\bm c$ fixed, we define
\begin{equation}
\label{eq:stepwise_sensitivity}
    G_k^{(n)} = \frac{ \left\|\widetilde{\bm a}_k^{(n)}-\bm a_k^{(n)}\right\|^2 }{ \sigma^2\left\|\bm\epsilon^{(n)}\right\|_F^2 }. 
\end{equation}
The score normalizes the action difference between the nominal and auxiliary actions by the squared displacement of the initial-noise perturbation.
Appendix~\ref{app:sensitivity_derivation} provides the connection between squared norm of the directional derivative of the action at offset $k$ for sufficiently small perturbations.

We use the resulting stepwise scores $\{G_k^{(n)}\}_{k=0}^{H-1}$ to guide joint nominal action chunk and its execution horizon selection.
The scores are computed from the paired outputs generated in Steps 1 and 2, without explicitly constructing Jacobians or training a separate evaluator.

\subsection{Joint Action Chunk and Execution Horizon Selection}
\label{sec:joint_selection}
\textbf{Step 4. Candidate filtering and horizon selection.} 
We filter nominal candidates by their initial-window sensitivity, then select the candidate with the longest execution horizon among those retained.
For each candidate $n$, define 
\begin{equation} 
\label{eq:initial_sensitivity} 
    S_n = \frac{1}{W}\sum_{k=0}^{W-1}G_k^{(n)}, 
\end{equation} 
where $W$ is the number of initial action timesteps included in the average, with $1\le W\le H$. 
The same averaging window is used for all candidates. 
We retain all candidates whose scores do not exceed the
median across the $N$ nominal candidates:
\begin{equation}
\label{eq:candidate_filtering}
    \mathcal K =
    \left\{
        n\in\{1,\ldots,N\} :
        S_n \le \operatorname{median}(S_1,\ldots,S_N)
    \right\}.
\end{equation}

For each retained candidate, we determine an execution horizon using its initial-window average $S_n$ as a sensitivity threshold.
Let $H$ denote the maximum execution length, we define
\begin{equation}
\label{eq:candidate_horizon}
\begin{aligned}
    h_n = \min\bigl(\{k\in\{W,\ldots,H-1\} :G_k^{(n)}>S_n\}\cup\{H\}\bigr)
\end{aligned}
\end{equation}
If $k$ is the first offset in $\{W,\ldots,H-1\}$ satisfying $G_k^{(n)}>S_n$, we set $h_n=k$. 
If no crossing occurs before the execution cap, $h_n=H$. 
The selected nominal chunk is $\bm A^{(n^\star)}$, and its execution horizon is $h^\star=h_{n^\star}$. 
We use the first $h^\star$ actions of the chunk, $[\bm a_0^{(n^\star)},\ldots, \bm a_{h^\star-1}^{(n^\star)}]$, as the control sequence. 
After executing the action sequence, we obtain an updated observation and repeat the candidate generation, assessment, and selection procedure. 
Each policy call thus provides $h^\star$ actions before replanning. 
\begin{algorithm}[h]
\caption{PICES: joint chunk and execution-horizon selection}
\label{alg:selection_overview}
\begin{algorithmic}[1]
    \State \textbf{Input:}
    Conditioning input $\bm c$, generator $V_\theta$,
    and parameters $N,W,\sigma$
    \State \textbf{Output:}
    Selected nominal chunk--horizon pair
    $(\bm A^{(n^\star)},h^\star)$

    \Statex
    \State \textit{// Candidate generation and sensitivity assessment}
    \State Independently sample
    $\bm Z^{(n)},\bm\epsilon^{(n)}\in\mathbb R^{d\times H}$
    with i.i.d. $\mathcal N(0,1)$ entries,
    $n=1,\ldots,N$
    \State $\widetilde{\bm Z}^{(n)}
    \gets \bm Z^{(n)}+\sigma\bm\epsilon^{(n)}$,
    \Comment{Perturb initial noise}
    \State Generate
    $\bm A^{(n)}=V_\theta(\bm c,\bm Z^{(n)})$
    and
    $\widetilde{\bm A}^{(n)}
    =V_\theta(\bm c,\widetilde{\bm Z}^{(n)})$
    for all $n$ in batched policy call

    \For{$n=1,\ldots,N$}
        \State Compute $G_k^{(n)}$ for $k=0,\ldots,H-1$
        using ~\eqref{eq:stepwise_sensitivity}
        \Comment{Stepwise sensitivity}
        \State $S_n
        \gets {W}^{-1}\sum_{k=0}^{W-1}G_k^{(n)}$
        \Comment{Initial-window sensitivity}
    \EndFor

    \Statex
    \State \textit{// Candidate filtering and horizon selection}
    \State $\mathcal K \gets
        \left\{
            n\in\{1,\ldots,N\} :
            S_n \le \operatorname{median}(S_1,\ldots,S_N)
        \right\}$
    \For{each $n\in\mathcal K$}
        \State $h_n\gets
        \min\bigl(
        \{k\in\{W,\ldots,H-1\}:G_k^{(n)}>S_n\}
        \cup\{H\}
        \bigr)$
        \Comment{Execution horizon selection}
    \EndFor

    \State $n^\star\in
    \operatorname*{arg\,max}_{n\in\mathcal K}h_n$,
    $h^\star\gets h_{n^\star}$
    \Comment{Action chunk selection}

    \State \textbf{Return:}
    $(\bm A^{(n^\star)},h^\star)$
\end{algorithmic}
\end{algorithm}



\begin{figure}[b]
    \centering
    \vspace{-2mm}
    \begin{minipage}[t]{0.42\linewidth}
        \vspace{0pt}
        \centering
        \includegraphics[width=0.95\linewidth]{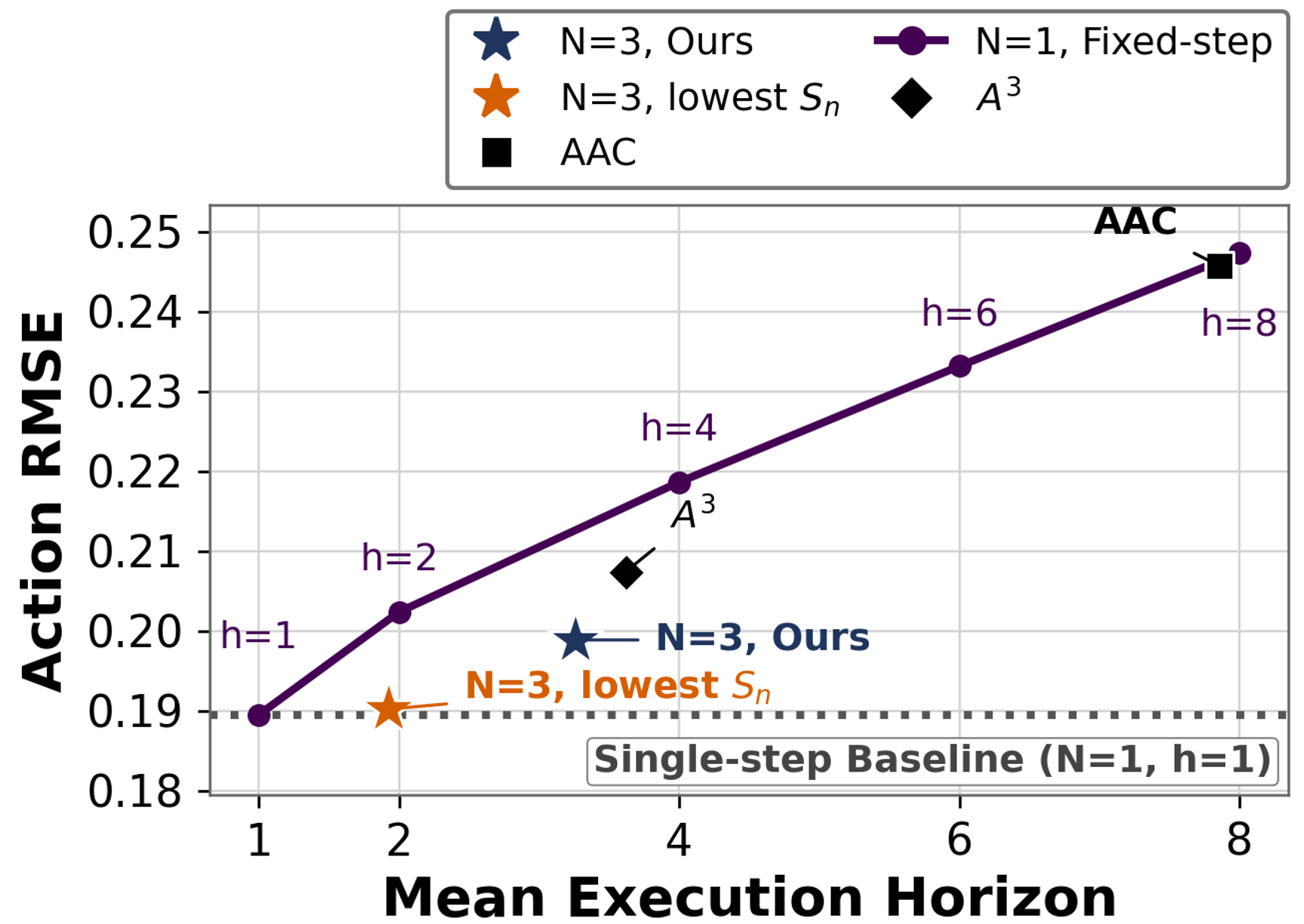}
    \end{minipage}\hfill
    \begin{minipage}[t]{0.55\linewidth}
        \vspace{0pt}
        \captionsetup{
            justification=raggedright,
            singlelinecheck=false,
            skip=0pt
        }
        \caption{
            \textbf{Open-loop evaluation of selected action horizons.}
            Action RMSE is computed between each selected nominal action and the corresponding reference action, following the evaluation protocol of the motivational study.
            Each fixed-step point is based on the single action chunk.
            For $A^3$, AAC, PICES, and the lowest-$S_n$ variant, the horizontal coordinate indicates the average of the horizons selected for individual method.
            PICES and the lowest-$S_n$ variant use $N=3$ nominal candidates;
            the latter selects the candidate with the smallest initial-window score ($S_n$).
        }
        \label{fig:our_rmse}
    \end{minipage}
    \vspace{-4mm}
\end{figure}

\vspace{-2mm}
\subsection{Open-loop Analysis of Chunk and Horizon Selection}
\label{sec:open_loop_analysis}
\paragraph{Effect of the candidate selection rule.}
We evaluate open-loop action prediction error across methods with fixed and adaptive execution horizons.
For each method, we compute RMSE between the first $h$ predicted actions and the corresponding recorded actions, following the evaluation protocol of the motivational study.
The fixed-step baseline uses a prescribed horizon with single action chunk. 
Figure~\ref{fig:our_rmse} reports action prediction RMSE against the average execution horizon.
The lowest-$S_i$ method obtains reference-action RMSE close to the single-step baseline while selecting approximately 1.9 actions on average. 
This observation supports the usefulness of sensitivity-guided candidate assessment for identifying action chunks with low reference-action error.
PICES retains candidates with low initial sensitivity and selects the candidate with the longest estimated execution horizon. 
The resulting average horizon increases to 3.3 actions, while the observed RMSE remains below that of the single-candidate, two-step execution horizon baseline.

\vspace{-2mm}

\section{Experiments}

\subsection{Experimental Setups}
\vspace{-1mm}

\textbf{Models and benchmarks.}
In simulation, we evaluate PICES with GR00T N1.7 on SimplerEnv–Fractal, $\pi_{0.5}$ and GR00T N1.5 on LIBERO. 
For each model–benchmark configuration, we retain the action dimensionality specified by the corresponding official repository and use default execution horizon for the fixed-horizon baseline.
We further evaluate $\pi_{0.5}$ on three real-world manipulation tasks using a UR10e robot arm. 

\textbf{Baselines and PICES configuration.}
We compare PICES with fixed-horizon execution, AAC, and $A^3$ based on each official repository. 
We also evaluate AAC$^{*}$ and $A^{3*}$, whose generation budgets match PICES’s total number of nominal and perturbed action chunks. 
Unless otherwise specified, PICES uses $N=3$ nominal candidates, perturbation scale of $\sigma=0.25$ and reference-window $W$ as default execution horizon specified by each model's official repository for the corresponding benchmark. 
Full model configurations, benchmark descriptions, evaluation protocols, and real-world task settings are provided in Appendix~\ref{app:exp_setup_benchmark}.

\begin{table}[!b]
  \centering
  \caption{
    \textbf{Main results on SimplerEnv--Fractal with GR00T N1.7.}
    We evaluate six Google Robot tasks with 100 episodes per task.
    We report task success rates (\%) and mean policy calls over successful episodes, together with per-call GPU latency and inference-time speedup on Jetson Orin and Thor.
    Speedup is relative to default fixed execution horizon.
    The best success rates and speedups are highlighted in bold.
  }
  
  \label{tab:simplerenv-sweep}
  \smallskip
  \small
  \setlength{\tabcolsep}{2.3pt}
  \renewcommand{\arraystretch}{1.05}

  \resizebox{\linewidth}{!}{%
    \begin{tabular}{@{}l*{12}{c}@{}}
      \toprule
      \multirow[c]{2}{*}[-12pt]{\textbf{Method}}
      & \multicolumn{7}{c}{\textbf{Success rate (\%) $\uparrow$}}
      & \multirow[c]{2}{*}[-12pt]{%
          \textbf{\makecell{Policy\\calls $\downarrow$}}}
      & \multicolumn{2}{c}{\textbf{Jetson Orin}}
      & \multicolumn{2}{c}{\textbf{Jetson Thor}} \\
      \cmidrule(lr){2-8}
      \cmidrule(lr){10-11}
      \cmidrule(lr){12-13}
      & \makecell{Pick\\Coke can}
      & \makecell{Pick\\object}
      & \makecell{Move\\near}
      & \makecell{Open\\drawer}
      & \makecell{Close\\drawer}
      & \makecell{Place in\\closed\\drawer}
      & \textbf{Average}
      &
      & \makecell{ Single call \\ (ms) $\downarrow$}
      & Speedup $\uparrow$
      & \makecell{ Single call \\ (ms) $\downarrow$}
      & Speedup $\uparrow$ \\
      \midrule

      Fixed ($h=1$)
      & \textbf{99} & \textbf{83} & \textbf{97} & 54 & 65 & 13 & 68.5
      & 46.8
      & \multirow{5}{*}{209.9}
      & 1.0 
      & \multirow{5}{*}{101.1}
      & 1.0 
      \\

      Fixed ($h=2$) & \textbf{99} & 80 & 92 & 57 & 67 & 15 & 68.3 & 26.6 & & 1.8 & & 1.8 \\

      Fixed ($h=4$)
      & 82 & 64 & 72 & 52 & 77 & \textbf{18} & 60.8
      & 19.4 & & 2.4 & & 2.4 \\

      Fixed ($h=6$) & 66 & 48 & 55 & 47 & 77 & 14 & 51.2 & 14.3 & & 3.3 & & 3.3 \\

      Fixed ($h=8$)
      & 65 & 42 & 44 & 35 & 78 & 8 & 45.3
      & 8.6 & & 5.4 & & 5.4 \\
      \midrule

      AAC
      & 58 & 42 & 56 & 43 & 76 & 9 & 47.3
      & 10.1 & 417.1 & 2.3 & 181.7 & 2.6 \\

      AAC$^{*}$
      & 58 & 36 & 64 & 48 & 77 & 17 & 50.0
      & 9.7 & 256.7 & \textbf{3.9}
      & 132.3 & \textbf{3.7} \\
        
      AutoHorizon
      & 67 & 47 & 62 & 44 & 74 & 8 & 50.3
      & 12.3 & 212.6 & 3.7 & 102.6 & \textbf{3.7} \\

      $A^{3}$
      & 79 & 56 & 71 & \textbf{56} & 75 & 13 & 58.3
      & 18.8 & 552.8 & 0.9 & 234.8 & 1.1 \\

      $A^{3*}$
      & 92 & 66 & 80 & 44 & 71 & 16 & 61.5
      & 20.4 & 398.1 & 1.2 & 192.2 & 1.2 \\
      \midrule
        
        
        \rowcolor{black!10}
        \textbf{PICES}
        & 97 & 82 & 91 & 53 & \textbf{79} & 10
        & \textbf{68.7}
        & 18.6 & 242.7 & 2.2 & 113.4 & 2.3 \\
        
        
        

      \rowcolor{black!10}
      \textbf{PICES}$^{**}$
      & \textbf{99} & 74 & 90 & 52 & \textbf{79} & 10 & 67.3
      & 14.3 & 242.7 & 2.8 & 113.4 & 2.9 \\
      \bottomrule
    \end{tabular}%
  }

  \par\vspace{2pt}
  \begin{minipage}{\linewidth}
    \raggedright
    \scriptsize
    \hfill $^{*}$Same total number of batched action samples as PICES, including nominal and perturbed chunks.\par
    \hfill $^{**}$PICES with $W=2$.\par
  \end{minipage}
\end{table}
\subsection{Quantitative Analyses}

\textbf{PICES improves task success while reducing total inference time.}
On SimplerEnv--Fractal with GR00T N1.7, PICES ($N=3$) achieves $68.7\%$ success with $60.3\%$ fewer policy calls, yielding inference-time speedups of $2.2\times$ on Jetson Orin and $2.2\times$ on Jetson Thor relative to the single-step baseline (Table~\ref{tab:simplerenv-sweep}).
The reported speedups account for each method's measured per-call GPU latency.
Increasing the fixed execution horizon to $h=8$, success rate falls to $45.3\%$, a decrease of $23.2\%$ from the single-step baseline.
Compared with the fixed-horizon baseline ($h=2$), PICES achieves inference-time speedups of $1.3\times$ on both Jetson Orin and Thor.
These results demonstrate PICES improves the trade-off between task success and inference time relative to the fixed-horizon baselines.

\textbf{Comparison with dynamic execution baselines.}
PICES also achieves higher task success than the competing adaptive execution methods. 
AAC and $A^3$ reduce the average number of policy calls
to $10.1$ and $18.8$, respectively. 
However, their success rates are $47.3\%$ and $58.3\%$, compared with $68.7\%$ for PICES ($N=3$).
The success-rate gap persists under matching sampling action budgets to PICES (including nominal and auxiliary chunks).
AAC$^{*}$ and $A^{3*}$ achieve success rates of $50.0\%$ and $61.5\%$, respectively, remaining $18.7\%$ and $7.2\%$  below PICES.

\begin{table}[!t]
  \centering
  \caption{
    \textbf{Results on LIBERO with $\pi_{0.5}$ and GR00T N1.5.}
    We report average success rates (SR, \%) and policy calls across the four LIBERO suites, together with per-call GPU latency and inference-time speedup on Jetson Orin and Thor.
  }
  \label{tab:libero-summary}
  \smallskip
  \small
  \setlength{\tabcolsep}{2.4pt}
  \renewcommand{\arraystretch}{1.08}

  \resizebox{\linewidth}{!}{%
    \begin{tabular}{@{}l*{6}{c}@{\hspace{8pt}}*{6}{c}@{}}
      \toprule
      & \multicolumn{6}{c}{\boldmath$\pi_{0.5}$}
      & \multicolumn{6}{c}{\textbf{GR00T N1.5}} \\
      \cmidrule(lr){2-7}
      \cmidrule(lr){8-13}

      \multirow{3}{*}{\textbf{Method}}
      & \multirow{3}{*}{\textbf{\makecell{Avg.\\SR\\(\%) $\uparrow$}}}
      & \multirow{3}{*}{\textbf{\makecell{Avg.\\Policy\\Calls $\downarrow$}}}
      & \multicolumn{2}{c}{\textbf{Jetson Orin}}
      & \multicolumn{2}{c}{\textbf{Jetson Thor}}
      & \multirow{3}{*}{\textbf{\makecell{Avg.\\SR\\(\%) $\uparrow$}}}
      & \multirow{3}{*}{\textbf{\makecell{Avg.\\Policy\\Calls $\downarrow$}}}
      & \multicolumn{2}{c}{\textbf{Jetson Orin}}
      & \multicolumn{2}{c}{\textbf{Jetson Thor}} \\
      \cmidrule(lr){4-5}
      \cmidrule(lr){6-7}
      \cmidrule(lr){10-11}
      \cmidrule(lr){12-13}

      & & &
      Single call & \multirow{2}{*}{Speedup $\uparrow$}
      & Single call & \multirow{2}{*}{Speedup $\uparrow$}
      & & &
      Single call & \multirow{2}{*}{Speedup $\uparrow$}
      & Single call & \multirow{2}{*}{Speedup $\uparrow$} \\

      & & &
      (ms) $\downarrow$ &
      & (ms) $\downarrow$ &
      & & &
      (ms) $\downarrow$ &
      & (ms) $\downarrow$ & \\
      \midrule

      Fixed ($h=5$)
      & 96.5 & 30.2 & 297.2 & 1.0 & 126.7 & 1.0
      & 92.9 & 30.1 & 149.3 & 1.0 & 62.8 & 1.0 \\
      \midrule

      AAC
      & 96.4 & 20.6 & 359.2 & 1.2 & 152.4 & 1.2
      & 93.3 & 24.2 & 282.5 & 0.7 & 107.2 & 0.7 \\

      AAC$^{*}$
      & 96.0 & 21.4 & 316.0 & 1.3 & 137.4 & 1.3
      & 92.8 & 25.8 & 185.9 & 0.9 & 77.7 & 0.9 \\

      AutoHorizon
      & 97.2 & 21.6 & 300.4 & 1.4 & 127.8 & 1.4
      & -- & -- & -- & -- & -- & -- \\

      $A^{3}$
      & 97.2 & \textbf{15.3} & 455.8 & 1.3 & 199.7 & 1.3
      & \textbf{94.9} & 29.7 & 277.7 & 0.5 & 124.9 & 0.5 \\

      $A^{3*}$
      & 96.0 & 29.2 & 421.6 & 0.7 & 187.3 & 0.7
      & 93.0 & 50.2 & 235.9 & 0.4 & 100.6 & 0.4 \\
      \midrule

      \rowcolor{black!10}
      \textbf{PICES}
      & 97.3 & 19.9 & 311.4 & 1.5 & 131.6 & 1.5
      & 94.5 & 21.4 & 178.5 & 1.2 & 74.8 & 1.2 \\

      \rowcolor{black!10}
      \textbf{PICES}$^{**}$
      & \textbf{97.5} & 16.4
      & 311.4 & \textbf{1.8} & 131.6 & \textbf{1.8}
      & 94.4 & \textbf{15.4}
      & 178.5 & \textbf{1.6} & 74.8 & \textbf{1.6} \\
      \bottomrule
    \end{tabular}%
  }

  \par\vspace{2pt}
  \begin{minipage}{\linewidth}
    \raggedright
    \scriptsize
    \hfill $^{*}$Same total number of batched action samples as PICES, including nominal and perturbed chunks.\par
    \hfill $^{**}$PICES with $W=8$. 
  \end{minipage}
\end{table}

\textbf{Evaluation on LIBERO.} 
On LIBERO, PICES improves success over Fixed ($h=5$) from 96.5\% to 97.3\% with $\pi_{0.5}$ and from 92.9\% to 94.5\% with GR00T N1.5 (Table~\ref{tab:libero-summary}).
Average policy calls decrease by 34.1\% and 28.9\%, respectively.
These reductions yield estimated inference-time speedups of $1.5\times$ on both devices with $\pi_{0.5}$, and $1.2\times$ with GR00T N1.5.
With $\pi_{0.5}$, $A^{3}$ achieves the lowest policy calls, but additional flow-matching computations for verification increase per-call latency 110~ms on Orin and 58~ms on Thor.
This overhead limits inference-time speedup to $1.3\times$ on both devices, below the $1.5\times$ achieved by PICES.
With GR00T N1.5 AAC achieves a slightly higher success rate than fixed-horizon baseline, but requires higher per-call latency, resulting in $0.7\times$ speedup on Orin and Thor. 
PICES improves success rate while achiving speedup on both devices.



\begin{table}[!b]
  \centering
  \caption{
    \textbf{Real-world task results using UR10e arm with $\pi_{0.5}$.}
    We report per-task and average success rates (SR, \%) and policy calls for three tasks, together with per-call GPU latency and inference-time speedup on Jetson Orin and Thor. Inference-time speedup is computed from the product of average policy calls and per-call GPU latency on each device, relative to Fixed ($h=8$).
  }
  
  \label{tab:real-robot}
  \smallskip
  \small
  \setlength{\tabcolsep}{2.6pt}
  \renewcommand{\arraystretch}{1.08}

  \resizebox{\columnwidth}{!}{%
    \begin{tabular}{@{}l*{12}{c}@{}}
      \toprule
      \multirow{3}{*}{\textbf{Method}}
      & \multicolumn{4}{c}{\textbf{Success rate (\%) $\uparrow$}}
      & \multicolumn{4}{c}{\textbf{Policy calls} $\downarrow$}
      & \multicolumn{2}{c}{\textbf{Jetson Orin}}
      & \multicolumn{2}{c}{\textbf{Jetson Thor}} \\
      \cmidrule(lr){2-5}
      \cmidrule(lr){6-9}
      \cmidrule(lr){10-11}
      \cmidrule(lr){12-13}

      & Tape & Stapler & Cube & \textbf{Average}
      & Tape & Stapler & Cube & \textbf{Average}
      & \makecell{Single call \\ (ms) $\downarrow$} & Speedup $\uparrow$
      & \makecell{Single call \\ (ms) $\downarrow$} & Speedup $\uparrow$ \\
      \midrule

        Fixed ($h=4$) & \textbf{97.2} & 77.8 & 58.3 & 77.8 & 614.8 & 236.8 & 485.9 & 445.8 & \multirow{4}{*}{298.5} & 0.39 & \multirow{4}{*}{126.7} & 0.39 \\
        
        Fixed ($h=8$)
        & 86.1 & \textbf{94.4} & 91.7 & 90.7 & 222.9 & 104.8 & 196.9 & 174.9 & & 1.00 & & 1.00 \\
        
        Fixed ($h=12$) & 88.9 & 80.6 & 88.9 & 86.1 & 120.8 & 74.4 & 111.1 & 102.1 & & 1.71 & & 1.71 \\
        
        Fixed ($h=16$) & 77.8 & 66.7 & 86.1 & 76.9 & 95.8 & 68.6 & 74.8 & 76.4 & & 2.29 & & 2.29 \\
        \midrule

        
        \rowcolor{black!10}
        \textbf{PICES}
        & 88.9 & \textbf{94.4} & \textbf{97.2} & \textbf{93.5} & 148.9 & 72.6 & 134.5 & 118.7
        & 314.7 & 1.39
        & 137.3 & 1.36 \\
        \bottomrule
    \end{tabular}%

  }
\end{table}

\textbf{Evaluation on Real-world tasks.} 
We evaluate PICES with $\pi_{0.5}$ on three manipulation tasks using a UR10e robot arm. 
Among the tested fixed-horizon baselines, $h=8$ achieves the highest average success rate of 90.7$\%$, as shown in Table~\ref{tab:real-robot}. 
We set the reference-window length of PICES to $W=8$ and use Fixed ($h=8$) as the baseline for comparisons. 
PICES increases the average success rate to 93.5$\%$, an improvement of 2.8 percentage points. 
PICES also reduces the average number of policy calls per episode from 174.9 to 118.7, a reduction of 32.1$\%$. 
Consequently, the reduction in policy calls yields the speedups of $1.39\times$ and $1.36\times$ on Orin and Thor, respectively, relative to Fixed ($h=8$).
These results demonstrate that the reduction at inference demand extends across the evaluated VLA backbones and manipulation benchmarks.


\subsection{Additional Analyses}
  \begin{wraptable}{r}{0.6\columnwidth}
    \vspace{-4mm}
    \centering
    \caption{
      \textbf{Effect of the number of nominal candidates on
      SimplerEnv--Fractal with GR00T N1.7.}
      We report average task success rates (\%) and average policy
      calls, together with per-call GPU latency and inference-time speedup relative to fixed execution horizon on Jetson Orin and Thor.
    }
    \label{tab:simplerenv-PICES}

    \smallskip
    \small
    \setlength{\tabcolsep}{2.3pt}
    \renewcommand{\arraystretch}{1.05}

    \resizebox{\linewidth}{!}{%
      \begin{tabular}{@{}l*{6}{c}@{}}
        \toprule
        \multirow{3}{*}{\textbf{Method}}
        & \multirow{3}{*}{\textbf{\makecell{Avg.\\SR\\(\%) $\uparrow$}}}
        & \multirow{3}{*}{\textbf{\makecell{Avg.\\Policy\\Calls $\downarrow$}}}
        & \multicolumn{2}{c}{\textbf{Jetson Orin}}
        & \multicolumn{2}{c}{\textbf{Jetson Thor}} \\
        
        \cmidrule(lr){4-5}
        \cmidrule(lr){6-7}
        
        & & &
        Single call & \multirow{2}{*}{Speedup $\uparrow$}
        & Single call  & \multirow{2}{*}{Speedup $\uparrow$} \\
        
        & & &
        (ms) $\downarrow$ &
        & (ms) $\downarrow$ & \\
        
        \midrule
        
        Fixed ($h=1$)
        & 68.5
        & 46.7
        & 209.8 & 1.0
        & 101.1 & 1.0 \\

        \textbf{PICES}, $N=1$
        & 67.3
        & 23.4
        & 214.3 & 2.0
        & 106.9 & 1.9 \\

        \rowcolor{black!10}
        \textbf{PICES}, $N=3$
        & 68.6
        & 18.5
        & 242.6 & \textbf{2.2}
        & 113.4 & \textbf{2.3} \\

        \textbf{PICES}, $N=5$
        & 68.5
        & 17.4
        & 303.6 & 1.9
        & 129.8 & 2.1 \\

        \textbf{PICES}, $N=7$
        & 68.0
        & 16.8
        & 357.7 & 1.6
        & 153.0 & 1.8 \\

        \textbf{PICES}, $N=9$
        & \textbf{69.0}
        & \textbf{16.1}
        & 397.8 & 1.5
        & 161.8 & 1.8 \\

        \bottomrule
      \end{tabular}%
    }
  \end{wraptable}
\paragraph{Effect of the number of nominal candidates.}
Table~\ref{tab:simplerenv-PICES} evaluates PICES with $N\in\{1,3,5,7,9\}$ on SimplerEnv--Fractal using GR00T N1.7.
Each policy call generates $2N$ chunks, comprising $N$ nominal and $N$ auxiliary chunks.
As $N$ increases from $1$ to $9$, the average number of policy calls per successful episode decreases from $23.4$ to $16.1$, a reduction of $31.1\%$.
On Orin, $N=3$ adds $5.1$\,ms per action sample generation.
At $N=9$, increased per action sample costs ($10.7$\,ms) outweigh further policy-call savings, consistent with the compute-bound behavior on Orin. 
Consequently, $N=3$ achieves $2.2\times$ speedup on Orin and $2.3\times$ on Thor.

\begin{table*}[!b]
  \centering
  \caption{
    \textbf{Compatibility of PICES with VLA acceleration methods
    on $\pi_{0.5}$.}
    We report success rates (\%) and average policy calls per episode across four LIBERO suites, together with per-call GPU latency and total inference-time speedup on Jetson Orin and Thor.
    Speedup is measured relative to Fixed ($h=5$) without visual token compression on the same device.
  }
  \label{tab:comp_vla_accel}
  \small
  \setlength{\tabcolsep}{3pt}
  \renewcommand{\arraystretch}{1.08}

  \resizebox{\textwidth}{!}{%
    \begin{tabular}{@{}l*{14}{c}@{}}
      \toprule
      \multirow[c]{2}{*}[-7.24pt]{\makecell{\textbf{Acceleration}\\\textbf{Method}}}
      & \multicolumn{5}{c}{\textbf{Success rate (\%) $\uparrow$}}
      & \multicolumn{5}{c}{\textbf{Policy calls $\downarrow$}}
      & \multicolumn{2}{c}{\textbf{Jetson Orin}}
      & \multicolumn{2}{c}{\textbf{Jetson Thor}} \\
      \cmidrule(lr){2-6}
      \cmidrule(lr){7-11}
      \cmidrule(lr){12-13}
      \cmidrule(lr){14-15}

      & Spatial & Goal & Object & Long & \textbf{Avg.}
      & Spatial & Goal & Object & Long & \textbf{Avg.}
      & \makecell{Single call \\ (ms) $\downarrow$ }
      & \makecell{Speedup $\uparrow$}
      & \makecell{Single call \\ (ms) $\downarrow $ }
      & \makecell{Speedup $\uparrow$} \\
      \midrule

      Baseline, Fixed
      & 99.0 & 97.4 & 97.2 & 92.2 & 96.5
      & 20.6 & 21.8 & 27.7 & 50.6 & 30.2
      & 297.2 & 1.0 & 126.7 & 1.0 \\

      \rowcolor{black!10}
      \textbf{PICES}
      & 99.0 & 98.4 & 99.6 & 92.2 & \textbf{97.3}
      & 13.9 & 14.4 & 18.0 & 33.4 & \textbf{19.9}
      & 311.4 & 1.5 & 131.6 & 1.5 \\
      \midrule

      FastV (40\%)
      & 99.0 & 97.4 & 98.0 & 89.6 & 96.0
      & 20.6 & 27.7 & 22.3 & 50.1 & 30.2
      & 237.5 & 1.3 & 106.3 & 1.2 \\

      \rowcolor{black!10}
      FastV (40\%) + \textbf{PICES}
      & 98.4 & 99.2 & 97.0 & 92.2 & 96.7
      & 14.0 & 18.0 & 14.7 & 33.5 & 20.1
      & 247.6 & 1.8 & 111.8 & 1.7 \\
      \midrule

      FastV (50\%)
      & 97.6 & 98.8 & 95.2 & 90.6 & 95.6
      & 21.2 & 27.6 & 22.3 & 51.2 & 30.6
      & 217.1 & 1.4 & 97.3 & 1.3 \\

      \rowcolor{black!10}
      FastV (50\%) + \textbf{PICES}
      & 98.0 & 98.2 & 95.4 & 92.0 & 95.9
      & 14.1 & 18.2 & 14.9 & 33.8 & 20.2
      & 228.5 & 1.9 & 100.8 & 1.9 \\
      \midrule

      GridS$_{16}$
      & 96.2 & 96.4 & 98.2 & 90.4 & 95.3
      & 21.4 & 23.1 & 28.4 & 52.7 & 31.1
      & 151.9 & 1.9 & 71.9 & 1.7 \\

      \rowcolor{black!10}
      GridS$_{16}$ + \textbf{PICES}
      & 98.2 & 98.8 & 96.2 & 92.4 & 96.4
      & 14.2 & 18.2 & 14.6 & 33.8 & 20.2
      & 160.5 & \textbf{2.8} & 75.4 & \textbf{2.5} \\
      \bottomrule
    \end{tabular}%
  }
\end{table*}

\paragraph{Compatibility with VLA Acceleration Methods}
\label{app:compat_vla_acceleration}

We evaluate PICES combined with FastV and GridS on LIBERO using $\pi_{0.5}$ with a fixed-horizon $h=5$ as the baseline. 
Both methods reduce per-call policy inference latency through visual token compression, using attention-based pruning for FastV and differentiable grid sampling for GridS.
For FastV, we evaluate pruning ratios of 40$\%$ and 50$\%$; For GridS, we use GridS$_{16}$, which samples 16 visual tokens per image. 

All speedups are relative to the fixed-horizon baseline without token pruning. 
Table~\ref{tab:comp_vla_accel} shows that PICES further reduces policy calls while maintaining or improving average success rates across all configurations. 
Combined with FastV (40$\%$), PICES achieves cumulative inference-time speedups of up to $1.8\times$ on Jetson Orin and $1.7\times$ on Jetson Thor; combined with GridS$_{16}$, the speedups reach $2.8\times$ and $2.5\times$, respectively. 
These results demonstrate that PICES provides additional efficiency gains on top of visual token compression.

\vspace{-2mm}
\subsection{Qualitative Analyses}

\begin{figure}[t]
    \centering
    \vspace{2mm}
    \begin{minipage}[t]{0.48\linewidth}
        \vspace{0pt}
        \centering
        \includegraphics[width=\linewidth]{
            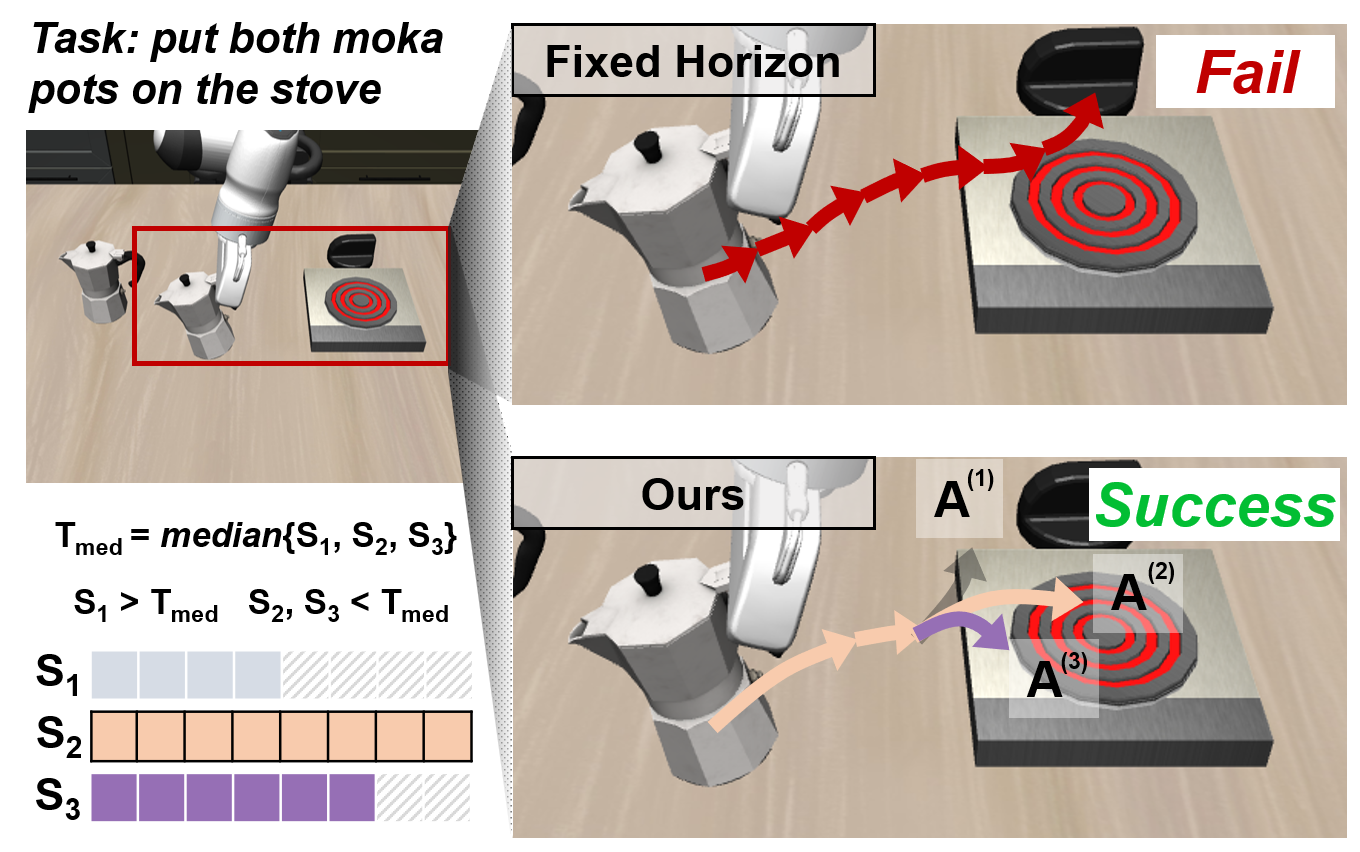
        }
        \vspace{-4mm}
        \caption{
            \textbf{Sensitivity-guided chunk selection in LIBERO--Long.}
            We compare Fixed ($h=5$, top) and PICES (bottom) using GR00T N1.5. 
            Colored arrows illustrate candidate motions, and the outlined bar identifies the selected execution horizon.
            PICES 
            selects $\bm A^{(2)}$ over $\bm A^{(3)}$ for the longer execution horizon ($h_2=10$).
        }
        \label{fig:qualitative_trajectory}
    \end{minipage}\hfill%
    \begin{minipage}[t]{0.48\linewidth}
        \vspace{0pt}
        \centering
        \vspace{-2mm}
        \includegraphics[width=\linewidth]{
            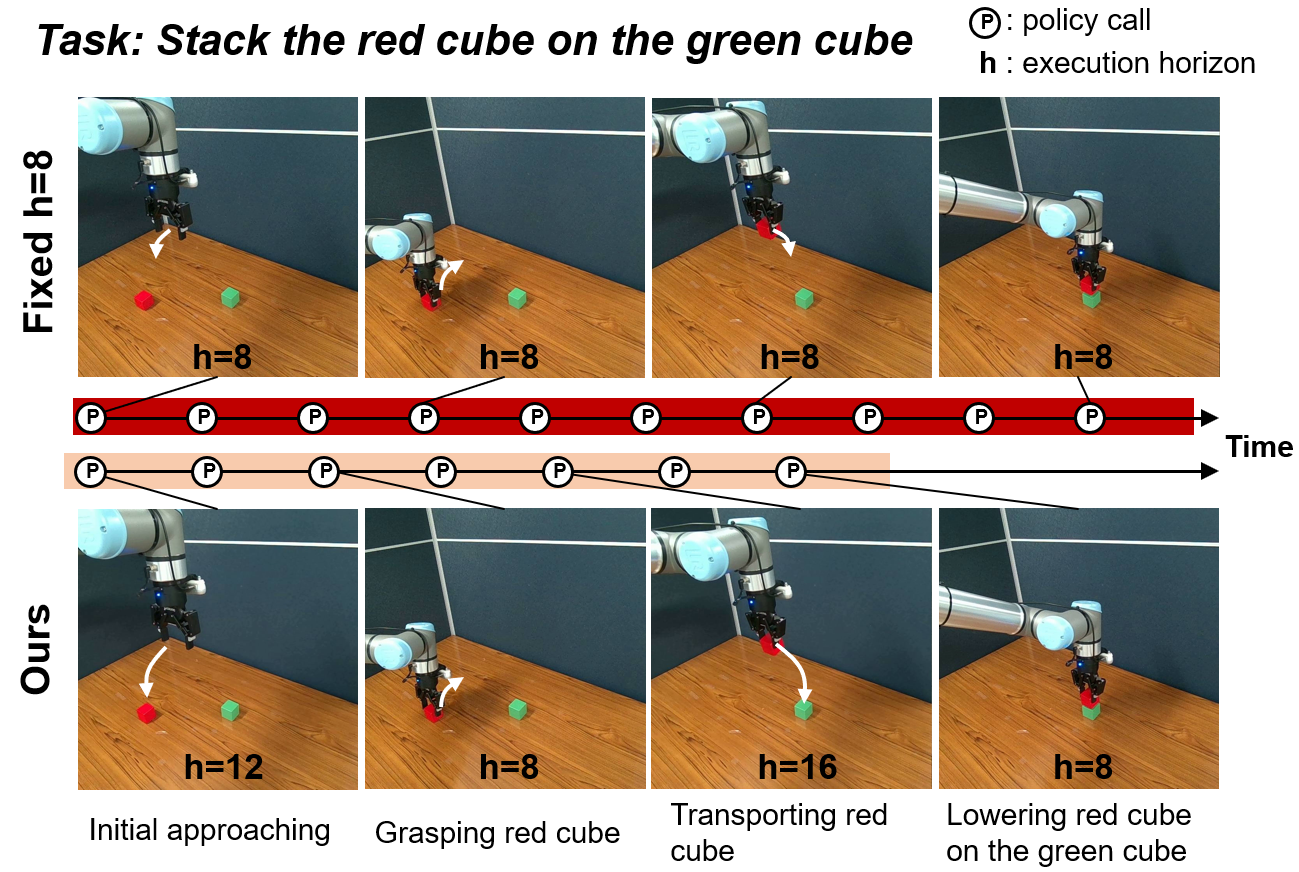
        }
        \vspace{-4mm}
        \caption{
            \textbf{Execution horizons across phases of real-world cube stacking.}
            We compare Fixed ($h=8$, top) and PICES (bottom) using $\pi_{0.5}$ on the UR10e.
            White arrows indicate robot motion, and circled P markers denote policy calls.
            Circled P markers indicate policy calls, and $h$ denotes the execution horizon.
        }
        \label{fig:qualitative_realworld}
    \end{minipage}
    \vspace{-4mm}
\end{figure}

\paragraph{Sensitivity-guided chunk selection from action candidates.}
Figure~\ref{fig:qualitative_trajectory} compares Fixed ($h=5$) and PICES using GR00T N1.5 on the LIBERO--Long task \textit{put both moka pots on the stove}.
The fixed-horizon baseline executes five actions from each sampled chunk, but the resulting motion fails to place the moka pot on the stove.
PICES evaluates three nominal chunks $N=3$ using the initial-window sensitivity scores, filtering out $\bm A^{(1)}$ with $S_1$ above the median.
Between $\bm A^{(2)}$ and $\bm A^{(3)}$, PICES selects $\bm A^{(2)}$ with longer execution horizon ($h_2=10$).
PICES executes the first ten actions of this nominal chunk, resulting in successful placement in the shown rollout.

\paragraph{Execution horizons across phases.}
Figure~\ref{fig:qualitative_realworld} compares fixed-horizon execution and PICES on \textit{stack the red cube on the green cube} using $\pi_{0.5}$ with the UR10e arm.
Given a prediction horizon of $H=16$, the baseline uses $h=8$ throughout execution, whereas PICES selects $h=12$ during the initial approach and $h=16$ while transporting the red cube.
During fine-grained phases such as near grasping and placement, PICES selects $h=8$, allowing replanning after fewer actions than coarse movement phases.
Adaptive horizon execution beyond the baseline allows reducing the policy calls to shorten task completion relative to the fixed horizon.

\vspace{-2mm}
\section{Conclusion}

We presented PICES, a training-free inference method for flow-matching VLA policies that jointly selects an action chunk and execution horizon.
PICES estimates stepwise action sensitivity by comparing outputs generated from each initial noise and locally perturbed counterpart. 
PICES exploits these scores to filter candidates, determine execution horizon, and select the nominal chunk with the longest execution horizon.
Experiments across multiple VLA backbones, simulation benchmarks, and real-world robot tasks show that PICES achieves higher task success rates while reducing cumulative inference time relative to the fixed-horizon baselines.


\newpage
\subsection*{AI use statement}

The research objectives, core ideas, and proposed method were developed by the authors without the use of generative AI tools. After independently formulating the research ideas, we used generative AI tools to help identify relevant prior work. We also used these tools to refine the manuscript’s language and clarity and to assist with code implementation.

The authors reviewed and verified all AI-generated outputs. In particular, AI-assisted code was reviewed and checked for correctness by at least two authors. We take full responsibility for the final content of this work, including all text, claims, code, and other artifacts produced with the assistance of generative AI.

\subsection*{Reproducibility statement}
The source code is available at \url{https://github.com/JHLee2314/PICES_VLA}.
The joint action chunk and execution horizon selection procedure is described in Section~\ref{sec:joint_selection}. 
Model configurations, baseline settings, and benchmark evaluation protocols are documented in Appendix~\ref{app:exp_setup_benchmark}. 
The open-loop evaluation procedure and reference-action error metric are detailed in Appendix~\ref{app:RMSE}. 
GPU profiling settings and latency measurements are provided in Appendix~\ref{app:AI_of_action}. 
The real-world platform, manipulation tasks, demonstration collection, policy fine-tuning, and evaluation protocol are described in Appendix~\ref{app:real_world_setup}.

\bibliographystyle{iclr-2027-style-files/iclr2027/iclr2027_conference.bst}
\bibliography{iclr2027}

\newpage

\appendix

\section{Horizon Distributions and Perturbation Sensitivity}
\begin{figure}[t]
    \centering
    \begin{minipage}[t]{0.64\linewidth}
        \vspace{0pt}
        \centering
        \includegraphics[width=\linewidth]{
            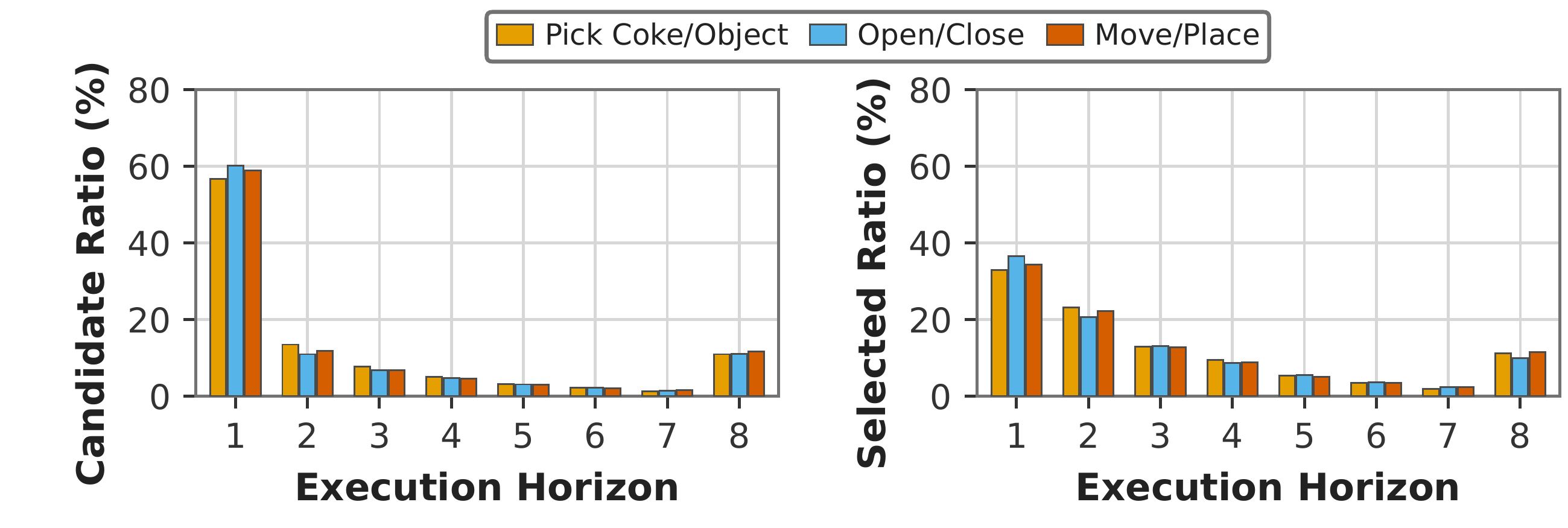
        }
        \vspace{-3mm}
        \caption{
        \textbf{Candidate and selected execution horizons in SimplerEnv ($N=3$).} \textbf{Left:} Execution horizons assigned to all nominal candidates. \textbf{Right:} Execution horizons of the nominal chunks selected by PICES at the same policy calls. 
        }
        \label{fig:simplerenv_horizons_exec}
    \end{minipage}\hfill%
    \begin{minipage}[t]{0.32\linewidth}
        \vspace{0pt}
        \centering
        \includegraphics[width=\linewidth]{
            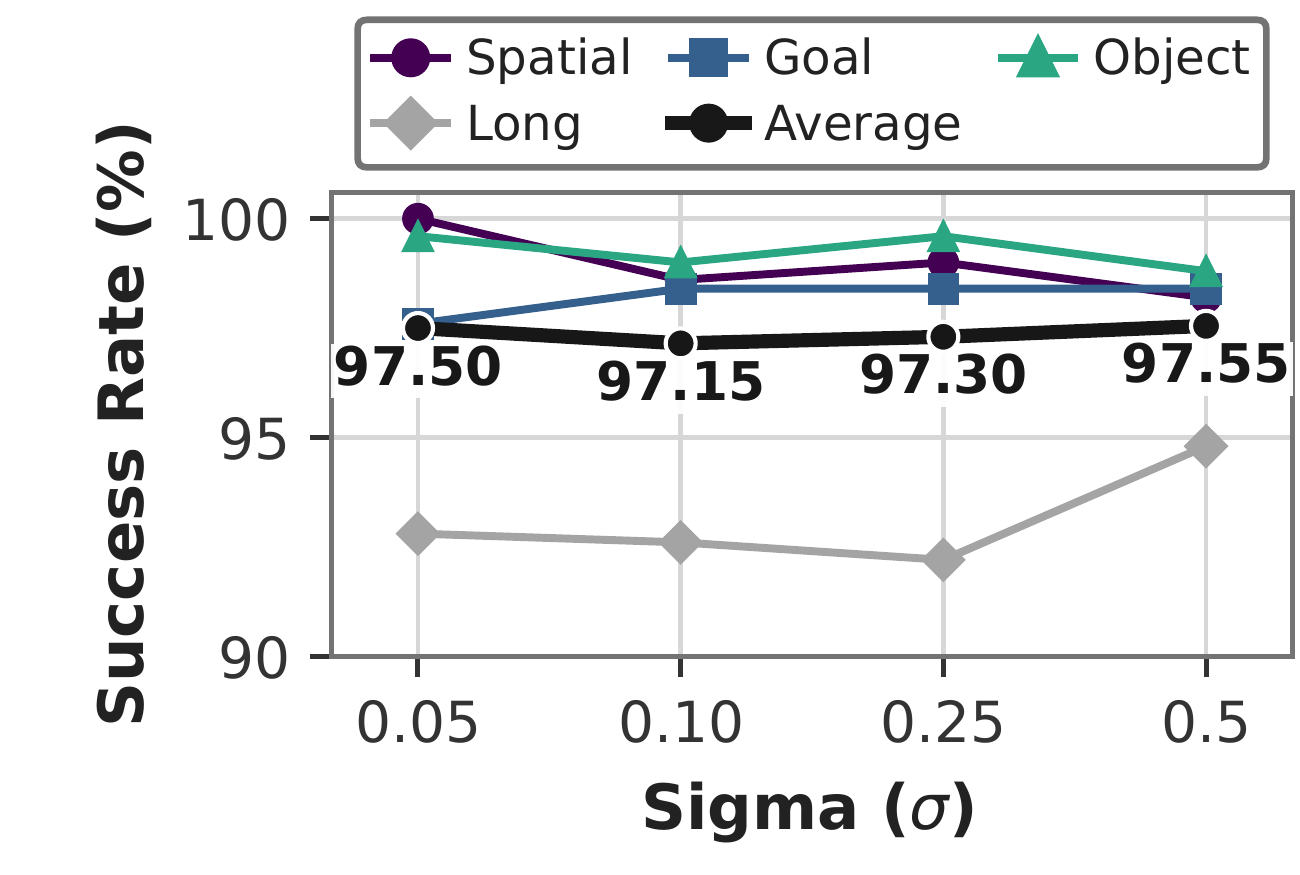
        }
        \vspace{-3mm}
        \caption{
            \textbf{Effect of the perturbation scale on LIBERO success rate.} We evaluate PICES at $\sigma\in\{0.05,0.10,0.25,0.50\}$. 
        }
        \label{fig:libero_sigma_success_rate}
    \end{minipage}
    \vspace{1mm}
\end{figure}

\paragraph{Distribution of execution horizons.}
To examine how PICES selects action chunks and execution horizons, we analyze the policy call records collected during closed-loop evaluation of GR00T N1.7 on SimplerEnv--Fractal with three $N=3$ nominal candidates. 
We group tasks by manipulation objective into three categories: Pick Coke/Object (Pick Coke Can and Pick Object), Open/Close (Open Drawer and Close Drawer), and Move/Place (Move Near and Place in Closed Drawer).

Figure~\ref{fig:simplerenv_horizons_exec} compares the execution horizons assigned to all nominal candidates with those of the chunks selected by PICES at the same recorded policy calls.
For this analysis, we apply the horizon-estimation rule to all three nominal candidates, including those excluded by the initial-sensitivity filter. 
Each call contributes three candidate horizons (left) and one selected horizon (right).
Across all groups, the selected distribution exhibits a lower proportion of one-step horizons and a higher proportion of intermediate horizons, particularly two to four steps.
The proportion of one-step horizons decreases from $59.07\%$ among nominal candidates to $34.93\%$ among PICES selections.
The corresponding average horizon increases from $2.49$ to $3.01$ actions.
The consistent shift across the three groups shows that candidate selection increases the execution length on average across different manipulation objectives.
The resulting distribution includes both short and long horizons, reflecting that PICES makes candidate-specific decisions during rollout. 


\paragraph{Effect of the perturbation scale.}
We examine the sensitivity of PICES's task performance to the perturbation scale $\sigma$ on LIBERO.
With $N=3$, we evaluate $\sigma\in\{0.05,0.10,0.25,0.50\}$ across the four task suites.
As shown in Figure~\ref{fig:libero_sigma_success_rate}, the average success rate across suites ranges from $97.15\%$ to $97.55\%$, a difference of $0.40$ percentage points.
Although individual suites exhibit variation, 
the average success rate changes little across the sigmas.

\newpage
\section{GPU Inference Cost Analysis}
\label{app:AI_of_action}
\paragraph{Operation Characteristics of Action Expert.}
Realizing benefits of multi-action chunks requires generating multiple candidate chunks without a proportional increase in inference latency in GPUs.
At the single batch, the arithmetic intensities (AI) of the $\pi_{0.5}$, GR00T N1.5, and GR00T N1.7 action experts are $59$, $63$, and $43$ FLOP/byte, respectively.
When AI remains smaller than ridge-point of each GPU, inference remains on the memory-bound condition.
The single-batch AI of VLA models lies in the memory-bound region of the roofline models for Jetson Orin and Thor, indicating that single-candidate generation does not fully utilize their compute throughput.

However, for the mulit-action chunk generation, $B$ initial-noise tensors are stacked along the batch dimension and processed by the action expert at each flow-matching step, avoiding repeated execution of the full VLA inference pipeline.
At a batch size of 6, the AI of $\pi_{0.5}$, GR00T N1.5, and GR00T N1.7 increases to 205.6, 190.6, and 182.4 FLOP/byte, respectively. 
Increasing the batch size to 8 further raises these values to 239.9, 222.1, and 223.0 FLOP/byte, exceeding the ridge point 207.5 FLOP/byte of Jetson Orin. 
Once the AI exceeds the ridge point, the operation turns into a compute-bound characteristic, where additional computation increases latency. 

Accordingly, for Jetson Orin, all three models approach the ridge point at a batch size of 6, without increasing latency due to additional computation. 
For Jetson Thor, the operations remain memory-bound even at a batch size of 20, indicating that the available compute throughput is still not fully utilized. 
However, beyond a batch size of 10, the growing cost of loading activations, in addition to the cost of loading action expert weights, leads to a smooth increase in latency.
Considering both hardware constraints and operation AI, these observations support setting the maximum batch size to 6 for Jetson Orin and 10 for Jetson Thor.

\subsection{Inference latency and batching overhead}
\label{app:GPU_latency}

\begin{table*}[!h]
  \centering
  \caption{
    \textbf{GPU latency across batch sizes on Jetson Orin and Jetson Thor.}
    We report the total VLA inference latency (ms) for batch sizes $B\in\{1,5,10,15,20\}$.
  }
  \label{tab:gpu-latency-batch-size}
  \smallskip
  \small
  \setlength{\tabcolsep}{4pt}
  \renewcommand{\arraystretch}{1.08}
  \resizebox{\textwidth}{!}{%
    \begin{tabular}{@{}l*{10}{r}@{}}
      \toprule
      \textbf{Model} & \multicolumn{5}{c}{\textbf{Jetson Orin}} & \multicolumn{5}{c}{\textbf{Jetson Thor}} \\
      \cmidrule(lr){2-6}\cmidrule(lr){7-11}
      & $B=1$ & $B=5$ & $B=10$ & $B=15$ & $B=20$ & $B=1$ & $B=5$ & $B=10$ & $B=15$ & $B=20$ \\
      \midrule
      $\pi_{0.5}$ & 310.4 & 361.6 & 464.1 & 581.1 & 676.6 & 127.2 & 149.8 & 179.3 & 215.3 & 267.0 \\
      GR00T N1.5 & 149.3 & 162.1 & 192.2 & 230.8 & 273.4 & 62.8 & 67.2 & 73.7 & 82.7 & 90.8 \\
      GR00T N1.7 & 209.9 & 235.4 & 301.6 & 361.6 & 401.1 & 101.1 & 111.5 & 124.5 & 148.3 & 166.2 \\
      \bottomrule
    \end{tabular}%
  }
\end{table*}


\begin{table*}[!h]
  \centering
  \caption{
    \textbf{Additional latency per batch across batch sizes on Jetson Orin and Jetson Thor.}
    We report per batch additional latency (ms) for batch sizes
    $B\in\{1,5,10,15,20\}$.
  }
  \label{tab:per-batch-gpu-latency}
  \smallskip
  \small
  \setlength{\tabcolsep}{4pt}
  \renewcommand{\arraystretch}{1.08}

  \resizebox{\textwidth}{!}{%
    \begin{tabular}{@{}l*{10}{r}@{}}
      \toprule
      \textbf{Model}
      & \multicolumn{5}{c}{\textbf{Jetson Orin}}
      & \multicolumn{5}{c}{\textbf{Jetson Thor}} \\
      \cmidrule(lr){2-6}
      \cmidrule(lr){7-11}
      & $B=1$ & $B=5$ & $B=10$ & $B=15$ & $B=20$
      & $B=1$ & $B=5$ & $B=10$ & $B=15$ & $B=20$ \\
      \midrule

      $\pi_{0.5}$
      & 9.03 & 10.25 & 15.37 & 18.05 & 18.31
      & 4.90 & 4.53 & 5.21 & 5.87 & 6.99 \\

      GR00T N1.5
      & 1.08 & 2.56 & 4.29 & 5.43 & 6.20
      & 0.96 & 0.87 & 1.08 & 1.32 & 1.40 \\

      GR00T N1.7
      & 2.91 & 5.10 & 9.17 & 10.11 & 9.56
      & 2.74 & 2.08 & 2.34 & 3.14 & 3.25 \\

      \bottomrule
    \end{tabular}%
  }
\end{table*}

\paragraph{Total inference latency and additional latency per batch.}
Table~\ref{tab:gpu-latency-batch-size} reports measured total inference latency. 
On Jetson Orin, increasing the batch size $B=1$ to $5$ results in latency increases of 3.5$\%$, 8.6$\%$, 12.1$\%$ for $\pi_{0.5}$, GR00T N1.5, and GR00T N1.7, respectively.
Batch size between $B=10$ to $20$, total latency increases by 10.9$\%$, 42.2$\%$, and 33.0$\%$ relative to $B=10$, respectively.
Table~\ref{tab:per-batch-gpu-latency} reports additional latency per batch, to quantify the cost of adding single action chunk.
Specifically, increasing the baseline batch size from $B=5$ to $B=10$ raises the incremental latency from $10.25$ to $15.37$ms for $\pi_{0.5}$, from $2.56$ to $4.29$ ms for GR00T N1.5, and from $5.10$ to $9.17$ms for GR00T N1.7.
Together with the roofline analysis, the measurements provide analytical and empirical support for the chosen generation budget.

\subsection{GPU Profiling Setup}

\begin{table}[!h]
  \centering
  \caption{GPU environments and evaluation settings.}
  \label{tab:gpu_setup}

  \small
  \setlength{\tabcolsep}{6pt}
  \renewcommand{\arraystretch}{1.05}

  \begin{tabular}{lcc}
    \hline
    \textbf{Setting}
    & \textbf{Jetson AGX Orin}
    & \textbf{Jetson AGX Thor} \\
    \hline

    Jetson Linux & R36.5.0 & R38.4.0 \\
    CUDA & 12.6 & 13.0 \\
    PyTorch & 2.8.0 & 2.13.0+cu130 \\
    TensorRT & 10.3.0.30 & 10.13.3.9 \\

    \hline

    Power mode & MAXN & MAXN \\
    Clock settings
    & \shortstack{Fixed GPU: 1300.5\,MHz\\Fixed EMC: 3199\,MHz}
    & \shortstack{Fixed GPU: 1575\,MHz\\Fixed EMC: 4266\,MHz} \\

    \hline

    Warm-up calls ($\pi_{0.5}$/GR00T) & 5/10 & 5/10 \\

    \hline
  \end{tabular}
\end{table}


\paragraph{Environment setup.}
VLA inference on edge device can become a latency bottleneck for responsive robot control, motivating inference acceleration. We therefore evaluate NVIDIA Jetson AGX Orin and Jetson AGX Thor as representative edge deployment platforms. Software environments and operating settings are listed in Table~\ref{tab:gpu_setup}.

\paragraph{Inference Configurations}
All models use \texttt{bfloat16} without quantization. For each model-device pair, PICES and all baselines, including AAC and $A^{3}$, are evaluated on the same optimized model implementation. The $\pi_{0.5}$ action expert uses device-tuned Triton kernels with CUDA graphs. 
GR00T System~1 uses the TensorRT deployment path provided by the official GR00T repository, while System~2 uses eager PyTorch with SDPA. 
We use 10 flow-matching steps for $\pi_{0.5}$ and 4 for both GR00T models. 
We evaluate $\pi_{0.5}$/GR00T inference latency after a 90\,s burn-in and 5/10 warm-up calls. 
Timing covers backbone inference and action generation from GPU-resident inputs, excluding input reprocessing, observation host-to-device transfer, environment interaction, and one-time initialization. PICES latency additionally includes probe construction and associated transfers, sensitivity computation, and candidate selection. 
Each PICES call generates $N$ nominal and $N$ perturbed chunks, with conditioning features computed once per call.

\newpage
\section{Open-loop Evaluation of Reference-Action Agreement}
\label{app:RMSE}

\paragraph{Evaluation contexts and action chunks.}
We evaluate open-loop action prediction error using the GR00T N1.7 SimplerEnv--Fractal checkpoint on the Fractal/RT-1 dataset.
Concretely, we uniformly sample 1,000 contexts without replacement from valid windows containing eight consecutive actions within a single episode.
Each context consists of the current image, robot state, and language instruction, together with the corresponding recorded action sequence.
The recorded actions serve as the reference action chunk for evaluation of one to eight action horizon.

\paragraph{Candidate sampling.}
We sample candidate chunks under identical conditioning inputs by drawing independent Gaussian initializations for each context.
For PICES and the lowest-$S_n$ variant, we generate one auxiliary chunk per candidate by adding Gaussian noise of scale $\sigma=0.25$ to its initial noise.
These auxiliary chunks provide sensitivity estimates, while the action execution uses nominal chunks.

\paragraph{Root-Mean-Square-Error to measure action error.}
Let $\hat{\bm a}_k$ and $\bm a_k$ denote the normalized prediction and reference vectors at action offset $k$.
For a prefix of length $h$, the error is
\begin{equation}
    \operatorname{RMSE}(h) =
        \sqrt{\frac{1}{dh} \sum_{k=0}^{h-1} \left\|\hat{\bm a}_k-\bm a_k\right\|^2 }
\label{eq:reference_prefix_rmse}
\end{equation}
where $d$ is the action dimension.
All action components and evaluated timesteps receive equal weight.
We report means of the per-evaluation RMSEs.

\paragraph{Oracle selected action chunk.}
The motivational study varies the number of nominal candidates from one to ten and evaluates fixed horizons of one to eight execution horizon.
For each context and candidate set, the oracle selects the candidate with the lowest RMSE at the specified horizon.
We then average these minimum errors to obtain the curves in Figure~\ref{fig:oracle_rmse}.
Selection is performed separately at each horizon, so the chosen candidate may change along a curve.
The oracle uses reference actions only for this analysis, and the dashed line marks the single-candidate, one-step baseline.

\paragraph{Fixed and adaptive horizon methods evaluation.}
The fixed-step baseline evaluates a single nominal chunk at a prescribed horizon.
PICES follows the selection procedure in Section~\ref{sec:joint_selection}, while the lowest-$S_n$ variant selects the candidate with the lowest initial sensitivity and applies the same horizon-estimation rule.
Both variants use an initial sensitivity window of one step.
For each adaptive method, we compute prefix RMSE at the horizon selected for each context, then average the errors and selected horizons separately.
Figure~\ref{fig:our_rmse} reports the resulting mean RMSE against the mean selected horizon.
We refer to Appendix~\ref{app:exp_setup_benchmark} for the AAC and $A^3$ configurations.

\newpage
\section{Interpretation of Stepwise Action Sensitivity}
\label{app:sensitivity_derivation}

\paragraph{Setup.}
We fix the conditioning input $\bm c$ and model parameters. 
Conditional on these fixed quantities, we sample
$\bm\epsilon^{(n)}$ with independent standard Gaussian entries.
Let $D=dH$ denote the dimension of the vectorized initial-noise matrix, and define
\begin{equation}
\label{eq:app_probe_direction}
\rho^{(n)}=\|\bm\epsilon^{(n)}\|_F,
\qquad
\bm u^{(n)}=
\frac{\operatorname{vec}(\bm\epsilon^{(n)})}{\rho^{(n)}},
\qquad
\delta^{(n)}=\sigma\rho^{(n)}.
\end{equation}
Here, $\bm u^{(n)}\in\mathbb R^D$ is the unit perturbation direction,
and $\delta^{(n)}$ is the perturbation magnitude. 
Let $\bm J_k^{(n)}\in\mathbb R^{d\times D}$ be the Jacobian of the action at offset $k$ with respect to the vectorized initial noise, evaluated at $\bm Z^{(n)}$.
This Jacobian differentiates the complete implemented action generator with respect to all initial-noise entries, including those associated with other action offsets. 
The index $k$ denotes a future action offset. 

\paragraph{Local directional interpretation.}
Normalizing output differences by input displacement provides a
finite-difference assessment of generator sensitivity, as used in
Jacobian Clamping~\citep{odena2018generator}.
Assume that the action-output map is differentiable at the nominal
initialization. For a fixed sampled probe, a first-order expansion as
$\sigma\to0$ gives
\begin{equation}
\label{eq:app_first_order}
\widetilde{\bm a}_k^{(n)}-\bm a_k^{(n)}
=\delta^{(n)}\bm J_k^{(n)}\bm u^{(n)}
+o\!\left(\delta^{(n)}\right).
\end{equation}
Such Taylor expansions are standard in analyses of small input
perturbations~\citep{bishop1995training}.
Substituting into Eq.~\eqref{eq:stepwise_sensitivity} yields
\begin{equation}
\label{eq:directional_sensitivity}
\begin{aligned}
G_k^{(n)}
&=\frac{
\|\widetilde{\bm a}_k^{(n)}-\bm a_k^{(n)}\|^2
}{(\delta^{(n)})^2}\\
&=\|\bm J_k^{(n)}\bm u^{(n)}\|^2+o(1).
\end{aligned}
\end{equation}
Thus, for sufficiently small realized perturbations, the score
approximates the squared norm of the directional derivative of the
action at offset $k$.

\paragraph{Interpretation at finite perturbation magnitude.}
For the finite perturbation used in generation, $G_k^{(n)}$ measures the normalized squared action change along the sampled displacement.
Its interpretation in terms of a directional derivative applies when the first-order approximation at the realized magnitude
$\delta^{(n)}=\sigma\|\bm\epsilon^{(n)}\|_F$.
The nonlinear remainder can affect the score at finite $\delta^{(n)}$, and normalization does not remove this effect.
All action offsets of a candidate share the same perturbation, so the profile $\{G_k^{(n)}\}_{k=0}^{H-1}$ describes how the corresponding actions respond to that perturbation.
These scores are computed from the paired outputs without explicitly constructing the Jacobian.
This local interpretation does not provide a bound on action prediction error. The usefulness of the profile for joint action-chunk and execution-horizon selection is evaluated empirically in Section~\ref{sec:open_loop_analysis}.

\newpage

\section{Detailed Experimental Setup and Benchmarks}
\label{app:exp_setup_benchmark}

\textbf{Models and implementation.} 
We evaluate PICES with three action-chunking VLA models: $\pi_{0.5}$, GR00T N1.5, and GR00T N1.7~\citep{intelligence2025pi_, bjorck2025gr00t}. 
We use their official implementations and apply PICES at inference time without additional training. 
Specifically, we evaluate GR00T N1.7 on SimplerEnv--Fractal~\citep{li2024evaluating}, $\pi_{0.5}$ and GR00T N1.5 on LIBERO~\citep{liu2023libero}. 
For each model–benchmark configuration, we retain the action dimensionality specified by the corresponding official repository and use its default execution horizon for the fixed-horizon baseline.

\textbf{Benchmarks and evaluation protocol.} 
We conduct simulation experiments on SimplerEnv and LIBERO. 
For SimplerEnv, we evaluate six manipulation tasks in the Fractal (Google Robot) setting, with 100 episodes per task. 
For LIBERO, we evaluate four task suites: LIBERO-Spatial, LIBERO-Goal, LIBERO-Object, and LIBERO-Long (LIBERO-10), with 10 tasks per suite and 50 evaluation episodes per task. 
\begin{itemize}
    \item \textbf{$\pi_{0.5}$ on LIBERO.}
    We use the released checkpoint and evaluation code provided by the official \texttt{openpi} repository, available at \url{https://github.com/Physical-Intelligence/openpi}.

    \item \textbf{GR00T N1.5 on LIBERO.}
    We use the released checkpoints for the four LIBERO suites and
    the evaluation scripts documented in the \texttt{n1.5-release}
    branch of the official \texttt{Isaac-GR00T} repository, available at
    \url{https://github.com/NVIDIA/Isaac-GR00T/tree/n1.5-release}.

    \item \textbf{GR00T N1.7 on SimplerEnv--Fractal.}
    We use the released Fractal checkpoint and evaluation scripts
    documented in the official \texttt{Isaac-GR00T} SimplerEnv example,
    available at
    \url{https://github.com/NVIDIA/Isaac-GR00T/blob/main}.
\end{itemize}

\textbf{Real-world setup }
\textcolor{black}{
    We evaluate our method across three manipulation tasks using a 6-DoF UR10e robot arm equipped with a Robotiq 2F-85 gripper: \texttt{Tape}, \texttt{Stapler}, and \texttt{Cube}.
    For each task, we collect 24 teleoperated demonstrations (72 in total) spanning varied initial object configurations.
    We evaluate each policy over 36 independent trials per task with randomized object placements.
    To assess both task effectiveness and computational efficiency, we report the Success Rate (\%) and the average number of policy calls per episode.
    See Appendix~\ref{app:real_world_setup} for the details.
}

\textbf{Baseline and PICES configurations.} 
We reproduce AAC, Autohorizon and $A^{3}$ using the code released in their respective repositories, retaining the action-sampling budgets specified by the released implementations. 
We additionally evaluate sample-matched variants that generate the same number of action samples per batch as PICES to align the GPU latency of batched action generation. 
These variants are denoted by AAC$^{*}$ and $A^{3*}$. The matched sampling budget includes all $N(1+1)$ nominal and perturbed action chunks generated by PICES.

Unless specified, PICES uses $N=3$ nominal candidates and a perturbation scale of $\sigma=0.25$. 
We use the default execution horizon specified by each model's official repository as the reference-window length $W$. 
We set the reference-window length $W$ to the default execution horizon specified in each model's official repository for the corresponding benchmark. 
For GR00T N1.5 on LIBERO, whose default execution horizon is one step, we use five action steps, matching the default execution horizon of $\pi_{0.5}$ on the same benchmark.
This window defines the initial sensitivity reference for candidate assessment, while PICES adaptively selects the execution horizon $h$ between $W$ and the prediction horizon $H$. 

\newpage

\section{Real-World Experimental Setup}
\label{app:real_world_setup}
\begin{figure}[!h]
    \centering
    \includegraphics[width=0.8\linewidth]{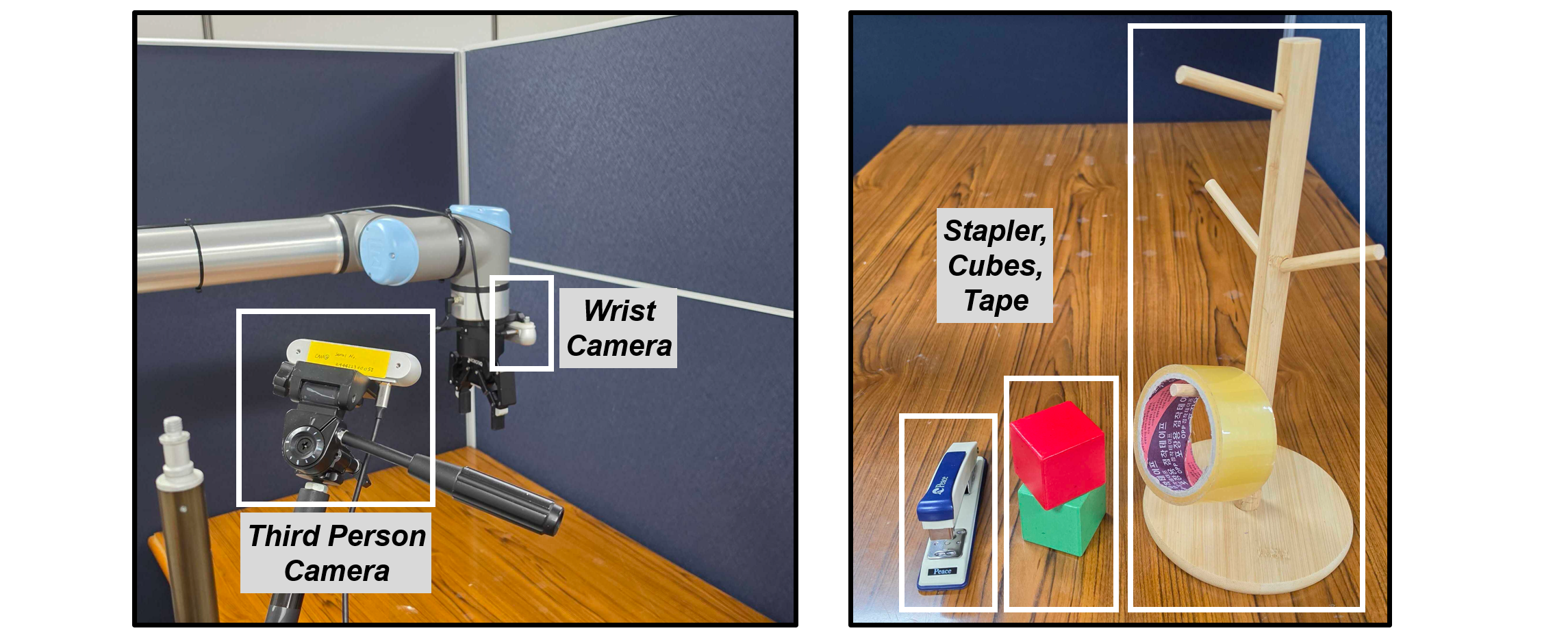}
    \caption{\textbf{Real-world experimental setup and task objects.} 
    Left: The UR10e robot with wrist-mounted and third-person cameras. Right: A tape and stand, a stapler, red and green cubes used for Tape Hanging, Stapler Pressing, and Cube Stacking, respectively.}
    \label{fig:realworld}
\end{figure}

\paragraph{Robot platform and control.}
As shown in Figure~\ref{fig:realworld}, our real-world experimental platform comprises a 6-DoF UR10e arm equipped with a Robotiq 2F-85 gripper. 
Visual observations are captured using RGB cameras streaming at 30\,Hz. 
The robot is operated in operational space at a control frequency of 30\,Hz, with actions parameterized as 7-dimensional vectors $[\Delta x, \Delta y, \Delta z, \Delta\text{roll}, \Delta\text{pitch}, \Delta\text{yaw}, g_t] \in \mathbb{R}^7$ representing end-effector Cartesian displacements, Euler angle rotations, and continuous gripper commands.

\begin{figure}[!h]
    \centering
    \includegraphics[width=\linewidth]{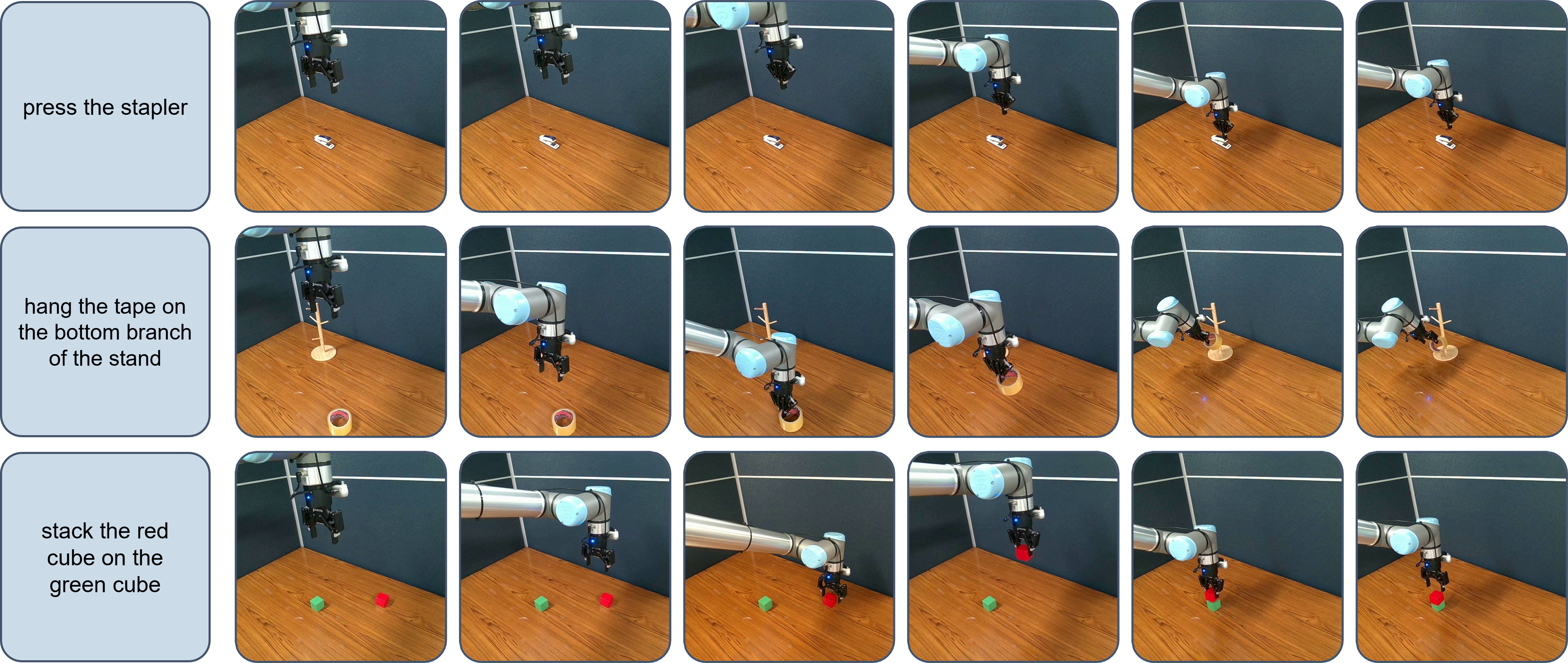}
    \vspace{-5mm}
    \caption{Task examples for real-world experiments}
    \label{fig:real_world_examples}
    \vspace{-3mm}
\end{figure}

\paragraph{Manipulation Tasks.}
We consider three contact-rich and fine-grained manipulation tasks: \textit{Tape Hanging}, \textit{Stapler Pressing}, and \textit{Cube Stacking}, which are abbreviated by their respective object names as \texttt{Tape}, \texttt{Stapler}, and \texttt{Cube} in the main text, as shown in Figure~\ref{fig:real_world_examples}.
\textit{Tape Hanging} requires picking up a tape from the table and hanging it onto the bottom branch of a multi-branch stand. 
\textit{Stapler Pressing} requires positioning the end-effector over a desktop stapler and applying downward force to depress the top lever and actuate the stapling mechanism. 
\textit{Cube Stacking} requires grasping the red cube, precisely stacking it atop the green cube, and retracting the end-effector upward with the gripper open, such that the tower remains stable without toppling.

\begin{table}[!h]
\fontsize{9}{11}\selectfont
\caption{Training hyperparameters and real-world experiment settings for fine-tuning $\pi_{0.5}$ on the UR10e platform.}
\label{tab:real_world_hyperparams}
\centering
\begin{tabularx}{\linewidth}{lX}
\toprule
\textbf{Hyperparameter / Setting} & \textbf{Value} \\
\midrule
Base checkpoint & $\pi_{0.5}$-base (OpenPI) \\
\# Training GPUs & 4 $\times$ NVIDIA RTX PRO 6000 Blackwell \\
Total batch size & 128 (32 per GPU) \\
\# Train steps & 5,000 steps ($\approx$ 17.7 epochs) \\
Fine-tuning mode & Full fine-tuning (all parameters) \\
Action chunk horizon & 16 steps \\
Control frequency & 30 Hz \\
Action representation & 7-DoF: $[\Delta x, \Delta y, \Delta z, \Delta\text{roll}, \Delta\text{pitch}, \Delta\text{yaw}, \text{gripper}]$ \\
Teleoperation interface & GELLO (collected at 30 Hz) \\
\# Demonstrations & 24 per task (72 total across \textit{Cube Stacking}, \textit{Tape Hanging}, \textit{Stapler Pressing}) \\
Evaluation protocol & 36 trials per task (108 rollouts total) under perturbed object poses \\
Evaluation metrics & Success Rate (\%), Average policy calls \\
\bottomrule
\end{tabularx}

\end{table}

\paragraph{Fine-tuning Configuration.}
All demonstration trajectories are recorded at 30\,Hz using a GELLO~\citep{wu2023gello}. 
For each task, we collect 24 teleoperated demonstrations (amounting to 72 trajectories in total) across diversified initial object configurations. 
We initialize our policy from the official $\pi_{0.5}\text{-base}$ checkpoint provided by OpenPI~\citep{openpi2025} and perform full-parameter fine-tuning on 4 $\times$ NVIDIA RTX PRO 6000 Blackwell GPUs. 
The model is trained for 5,000 steps with an effective batch size of 128 (32 per GPU), corresponding to approximately 17.7 training epochs, using an action chunk prediction horizon of $H = 16$. 

\paragraph{Evaluation Protocol.}
Each policy is evaluated on 36 independent trials per task (108 rollouts total) with 12 distinct initial object placements. 
We report the task success rate (\%) and the average number of policy calls per episode to measure execution reliability and inference efficiency. 
Detailed training and deployment configurations are summarized in Table~\ref{tab:real_world_hyperparams}.

\newpage
\section{Additional Related Work}
\paragraph{Action Chunking in VLA Models.}
\label{app:related_action_chunk}

Action chunking policy predicts a sequence of future actions conditioned on current observation, improving temporal consistency. 
Recent VLAs such as CogACT, $\pi$, and GR00T series use transformer-based action experts, implemented as Diffusion Transformers (DiTs), to model multimodal distributions over continuous action chunks~\citep{li2024cogact,black2024pi_0,bjorck2025gr00t}. 
At inference time, such policies execute only a fixed-length horizon of each predicted chunk. 
Executing a long fixed-length actions reduces policy calls but limits the incorporation of new observations, motivating adaptive execution horizons~\citep{liang2026adaptive, wang2026vla, chen2026dynamic, zhang2026hipolicy}.
MoH requires auxiliary module for multiple horizons and uses cross-horizon agreement to determine an executable actions~\citep{jing2025mixture}.
DEHP uses a lightweight predictor trained with online reinforcement learning to adapt the execution horizon of a frozen pretrained action-chunking policy~\citep{zhao2026dynamic}.
AAC estimates a single execution horizon from cross-sample action entropy~\citep{liang2026adaptive}.
AutoHorizon estimates an execution horizon for each predicted chunk using the self-attention patterns~\citep{wang2026vla}.
$A^{3}$ selects a representative draft from sampled trajectories and determines execution horizon through an additional batched verification pass~\citep{chen2026dynamic}.
In contrast, our method instead estimates a horizon for each candidate and uses the estimated execution length to jointly choose the trajectory and its execution extent in inferece time without training.

\paragraph{Acceleration for vision-language-action models}
\label{app:related_accel}
Recent works on efficient VLA inference seek to reduce cumulative inference time using various methods.
SmolVLA overlaps action prediction with execution and evaluates efficiency using task-completion time and the number of successful pick-and-place operations within a fixed time budget~\citep{shukor2025smolvla}.
To reduce inference overhead, prior work exploits redundancy in visual inputs and model computation through token pruning, channel pruning, and model scheduling~\citep{wang2025specprune,li2025sp,liu2025vla,huang2025dadu, huang2026environment, feng2026see}.
FastV, originally developed for vision-language models, prunes visual tokens using attention scores from early layers to reduce computation in subsequent layers~\citep{chen2024image}.
VLA-Cache exploits temporal redundancy by reusing cached key-value representations of minimally changed visual tokens across adjacent observations~\citep{xu2026vla}.
GridS reduces the computational cost of vision-language-action models by learning task-aware sampling coordinates and extracting pruned visual tokens via differentiable bilinear interpolation~\citep{feng2026see}.
Our method is orthogonal to token pruning, which reduces computation within individual policy calls.
DaDu-Corki predicts future trajectories to reduce expensive model invocations~\citep{huang2025dadu}.
PICES examines how candidate-specific horizon selection can reduce policy calls, while explicitly accounting for the additional cost of generating and assessing multiple action candidates. 

\newpage
\section{Detailed Simulation Results on LIBERO}

\begin{table}[!h]
  \centering
  \caption{
    \textbf{LIBERO results with $\pi_{0.5}$ and GR00T N1.5.}
    We report per-suite and average success rates (\%) and policy calls,
    together with single-call GPU latency and inference-time speedup on
    Jetson Orin and Thor. Fixed ($h=x$) executes $x$ actions per policy call;
    single-call latency is shared across the five fixed-horizon settings for
    each model and device. Speedups are reported relative to Fixed ($h=5$).
  }
  \label{tab:libero_appendix_results}
  \smallskip
  \small
  \setlength{\tabcolsep}{2.6pt}
  \renewcommand{\arraystretch}{1.08}

  \resizebox{\columnwidth}{!}{%
    \begin{tabular}{@{}l*{14}{c}@{}}
      \toprule
      \multirow{2}{*}{\textbf{Method}}
      & \multicolumn{5}{c}{\textbf{Success rate (\%) $\uparrow$}}
      & \multicolumn{5}{c}{\textbf{Policy calls} $\downarrow$}
      & \multicolumn{2}{c}{\textbf{Jetson Orin}}
      & \multicolumn{2}{c}{\textbf{Jetson Thor}} \\
      \cmidrule(lr){2-6}
      \cmidrule(lr){7-11}
      \cmidrule(lr){12-13}
      \cmidrule(lr){14-15}

      & Spatial & Goal & Object & Long & \textbf{Average}
      & Spatial & Goal & Object & Long & \textbf{Average}
      & \makecell{Single call \\ (ms) $\downarrow$} & Speedup $\uparrow$
      & \makecell{Single call \\ (ms) $\downarrow$} & Speedup $\uparrow$ \\
      \midrule
      \multicolumn{15}{c}{\boldmath$\pi_{0.5}$} \\
      \midrule
      Fixed ($h=1$)
      & 99.0 & 92.8 & 91.4 & 84.4 & 91.9
      & 102.8 & 110.6 & 145.7 & 272.9 & 158.0
      & \multirow{5}{*}{297.2} & 0.2 & \multirow{5}{*}{126.7} & 0.2 \\
      Fixed ($h=2$)
      & 99.0 & 95.4 & 94.8 & 86.6 & 94.0
      & 51.2 & 55.3 & 71.7 & 132.1 & 77.6
      & & 0.4 & & 0.4 \\
      Fixed ($h=5$)
      & 98.0 & 97.6 & 98.2 & 92.0 & 96.5
      & 20.6 & 21.8 & 27.7 & 50.6 & 30.2
      & & 1.0 & & 1.0 \\
      Fixed ($h=8$)
      & \textbf{99.4} & 98.6 & 98.2 & 93.4 & 97.4
      & 13.1 & 13.7 & 17.2 & 31.5 & 18.9
      & & 1.6 & & 1.6 \\
      Fixed ($h=10$)
      & 98.2 & 98.2 & 97.6 & 93.8 & 97.0
      & 10.5 & 11.1 & 13.7 & 25.6 & \textbf{15.2}
      & & \textbf{2.0} & & \textbf{2.0} \\
      \midrule
      AAC
      & 97.6 & 94.2 & 98.8 & 94.8 & 96.4
      & 15.5 & 15.6 & 18.8 & 32.5 & 20.6
      & 359.2 & 1.2 & 152.4 & 1.2 \\
      AAC$^{*}$
      & 97.4 & 94.2 & 99.0 & 93.2 & 96.0
      & 16.3 & 16.3 & 19.8 & 33.7 & 21.4
      & 316.0 & 1.3 & 137.4 & 1.3 \\
      AutoHorizon
      & 98.6 & 98.2 & 97.0 & \textbf{95.2} & 97.2
      & 15.2 & 19.5 & 16.0 & 35.6 & 21.6
      & 300.4 & 1.4 & 127.8 & 1.4 \\
      $A^{3}$
      & 99.0 & \textbf{99.0} & 97.0 & 93.8 & 97.2
      & 10.6 & 11.2 & 13.9 & 25.5 & 15.3
      & 455.8 & 1.3 & 199.7 & 1.3 \\
      $A^{3}$$^{*}$
      & 98.8 & 97.4 & 97.6 & 90.4 & 96.0
      & 19.7 & 20.8 & 26.9 & 49.3 & 29.2
      & 421.6 & 0.7 & 187.3 & 0.7 \\
      \rowcolor{black!10}
      \textbf{PICES, \boldmath$N=3$}
      & 99.0 & 98.4 & \textbf{99.6} & 92.2 & 97.3
      & 13.9 & 14.4 & 18.0 & 33.4 & 19.9
      & 311.4 & 1.5 & 131.6 & 1.5 \\
      \rowcolor{black!10}
      \textbf{PICES$^{**}$, \boldmath$N=3$}
      & 98.4 & 98.6 & 99.4 & 93.6 & \textbf{97.5}
      & 11.5 & 11.9 & 14.8 & 27.4 & 16.4
      & 311.4 & 1.8 & 131.6 & 1.8 \\
      \midrule
      \multicolumn{15}{c}{\textbf{GR00T N1.5}} \\
      \midrule
      Fixed ($h=1$)
      & 89.6 & 90.0 & 92.6 & 71.4 & 85.9
      & 108.6 & 115.3 & 140.7 & 259.1 & 155.9
      & \multirow{5}{*}{149.3} & 0.2 & \multirow{5}{*}{62.8} & 0.2 \\
      Fixed ($h=2$)
      & 91.4 & 91.0 & 93.8 & 78.0 & 88.6
      & 53.7 & 56.7 & 69.6 & 129.0 & 77.2
      & & 0.4 & & 0.4 \\
      Fixed ($h=5$)
      & 94.8 & 95.0 & 95.8 & 85.8 & 92.9
      & 21.3 & 22.5 & 27.9 & 50.5 & 30.6
      & & 1.0 & & 1.0 \\
      Fixed ($h=8$)
      & 95.2 & \textbf{95.8} & 95.6 & 86.4 & 93.3
      & 13.5 & 14.3 & 17.5 & 34.2 & 19.9
      & & 1.5 & & 1.6 \\
      Fixed ($h=16$)
      & 94.0 & 93.8 & 97.0 & 86.0 & 92.7
      & 7.0 & 7.3 & 9.0 & 16.2 & \textbf{9.9}
      & & \textbf{3.1} & & \textbf{2.0} \\
      \midrule
      AAC
      & 94.0 & 92.8 & 97.8 & 88.6 & 93.3
      & 18.6 & 15.7 & 25.1 & 37.9 & 24.2
      & 282.5 & 0.7 & 107.2 & 0.7 \\
      AAC$^{*}$
      & 94.0 & 92.0 & 96.6 & 88.6 & 92.8
      & 19.7 & 17.0 & 26.5 & 39.9 & 25.8
      & 185.9 & 0.9 & 77.7 & 0.9 \\
      $A^{3}$
      & 94.6 & 95.6 & \textbf{99.6} & \textbf{90.0} & \textbf{94.9}
      & 21.2 & 19.4 & 27.7 & 50.7 & 29.7
      & 277.7 & 0.5 & 124.9 & 0.5 \\
      $A^{3}$$^{*}$
      & 94.8 & 93.0 & 94.5 & 89.6 & 93.0
      & 35.3 & 33.1 & 46.4 & 85.9 & 50.2
      & 235.9 & 0.4 & 100.6 & 0.4 \\
      \rowcolor{black!10}
      \textbf{PICES, \boldmath$N=3$}
      & \textbf{97.4} & 95.6 & 95.6 & 89.4 & 94.5
      & 15.7 & 15.5 & 19.6 & 36.3 & 21.4
      & 178.5 & 1.2 & 74.8 & 1.2 \\
      \rowcolor{black!10}
      \textbf{PICES$^{**}$, \boldmath$N=3$}
      & 95.2 & 94.8 & 97.6 & \textbf{90.0} & 94.4
      & 10.8 & 11.1 & 14.0 & 25.8 & 15.4
      & 178.5 & 1.6 & 74.8 & 1.6 \\
      \bottomrule
    \end{tabular}%
  }
  \par\vspace{2pt}
  \begin{minipage}{\linewidth}
    \raggedleft
    \scriptsize
    $^{*}$Same number of action samples as PICES.\\
    $^{**}$PICES with $W=8$.
  \end{minipage}
  \vspace{-2mm}
\end{table}

\end{document}